%% file: main.tex
\documentclass[10pt]{article} % For LaTeX2e
\usepackage[preprint]{tmlr}

\input{math_commands.tex}

\usepackage{hyperref}
\usepackage{url}

\def\vx{\mathbf{x}}
\def\vk{\mathbf{k}}
\def\vq{\mathbf{q}}
\def\vp{\mathbf{p}}
\def\va{\mathbf{a}}

\def\vs{\mathbf{s}}

\usepackage{algpseudocode}
\newcommand{\tsb}{\textsubscript}
\newcommand{\ta}[1]{{ #1 }}

\usepackage{eqparbox}
\usepackage{comment}
\usepackage{multirow}
\usepackage{makecell}
\usepackage{graphicx}
\usepackage{booktabs}

\usepackage{amsmath}
\usepackage{amssymb}
\usepackage[ruled,vlined,linesnumbered]{algorithm2e}
\usepackage{bm}

\title{GAP-Prompt: Gated Adaptive Prompting for Efficient Continual Learning}

\author{\name Trung-Anh Dang \email trung-anh.dang@univ-orleans.fr \\
      \addr LIFO UR 4022 \\
      Université d'Orléans, INSA CVL, Orléans, 45067, France
      \AND
      Duy-Cuong Bui \email duy-cuong.bui@ensea.fr \\
      \addr ETIS \\
      CY Cergy Paris University, ENSEA, CNRS, Cergy, 95000, France
      \AND
      \name Ngoc-Son Vu \email son.vu@utt.fr\\
      \addr LIST3N \\
      Université de Technologie de Troyes, Troyes, 10010, France
      \AND
      \name Christel Vrain \email christel.vrain@univ-orleans.fr \\
      \addr LIFO UR 4022 \\
      Université d'Orléans, INSA CVL, Orléans, 45067, France
      \AND
      \name Vincent Nguyen \email vincent.nguyen@univ-orleans.fr \\
      \addr LIFO UR 4022 \\
      Université d'Orléans, INSA CVL, Orléans, 45067, France
}

\def\month{MM}  % Insert correct month for camera-ready version
\def\year{YYYY} % Insert correct year for camera-ready version
\def\openreview{\url{https://openreview.net/forum?id=XXXX}} % Insert correct link to OpenReview for camera-ready version

\begin{document}

\maketitle

\begin{abstract}
Continual learning faces the persistent challenge of catastrophic forgetting, where sequential task updates degrade previously acquired knowledge. While prompt-based methods integrated with pre-trained models offer a compelling solution by freezing the backbone, they often rely on static, task-level prompting strategies that overlook fine-grained intra-task diversity. In this paper, we propose Gated Adaptive Prompting (GAP-Prompt), a novel method that introduces instance-level adaptability to the prompting process. GAP-Prompt consists of three synergistic modules: (1) instance-conditioned gating, which dynamically determines optimal prompt injection layers for each individual image; (2) dynamic knowledge fusion, which performs instance-aware aggregation of current and historical prompts, enabling knowledge integration across tasks; and (3) shared prompt distillation, which anchors foundational knowledge in early shared layers to mitigate forgetting. Extensive evaluations on CIFAR-100, ImageNet-R, and CUB-200 benchmarks demonstrate that GAP-Prompt consistently achieves state-of-the-art performance. Notably, on the fine-grained CUB-200 dataset, GAP-Prompt reaches \text{87.29\%} accuracy, approaching the joint training upper bound (88.00\%) and outperforming existing methods by a significant margin.
\end{abstract}

%-------------------------------------------------------------------------
\section{Introduction}
\label{sec:intro}

Continual learning (CL) seeks to develop autonomous systems capable of acquiring knowledge from a sequential stream of tasks, mimicking the lifelong learning process of biological systems. However, standard neural networks struggle with catastrophic forgetting \citep{mccloskey_1989, robins1995_catastrophic_fr}, where the weight updates required for new tasks disrupt previously learned knowledge. 
% where the weight updates required for new tasks lead to the abrupt erosion of previously learned representations. 
Traditional mitigation strategies, such as regularization-based methods \citep{friedemann_si_2017, fini2022cassle, sangwon_neurips_2020} and experience replay \citep{buzzega2020darkexperiencegeneralcontinual, madaan_2022_lump, co2l}, often encounter a fundamental trade-off between memory efficiency, data privacy, and computational overhead as the number of tasks increases.

Recently, the paradigm of prompt-based CL has emerged as a compelling solution, particularly when integrated with large-scale pre-trained models like the Vision Transformer (ViT)~\citep{dosovitskiy2020image}. Building on visual prompt tuning~\citep{jia2022visual}, prompt-based CL keeps the backbone model's weights frozen to preserve foundational knowledge. Instead, it optimizes a small set of task-specific parameters known as \textbf{prompts}. These prompts serve as dynamic instructions that guide the pre-trained model to adapt to novel tasks without altering its core parameters~\citep{wang2022learning, wang2022dualprompt, smith2023coda}. Consequently, this approach significantly alleviates catastrophic forgetting while ensuring high scalability in complex, real-world learning environments. 
%VN: However, 
Existing prompt-based CL methods typically adopt one of two strategies for managing prompts across tasks. Some approaches rely on a shared prompt pool \citep{wang2022learning, smith2023coda} that is reused across tasks, which can lead to interference as new tasks overwrite previously useful prompts. In contrast, other methods learn task-specific prompts independently \citep{wang2022dualprompt, wang2023spromptslearningpretrainedtransformers}, limiting effective knowledge reuse from previously learned prompts. These limitations highlight the need for mechanisms that can dynamically combine useful prompt knowledge across tasks.

\begin{figure}[ht]
    \centering
    \includegraphics[width=1.0\linewidth]{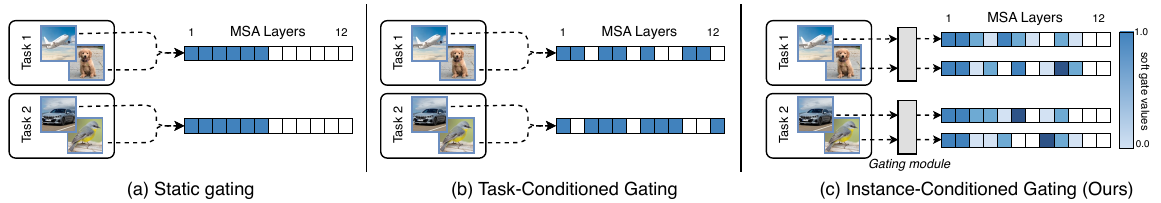}
    \caption{Conceptual comparison of different prompt gating mechanisms, i.e., determining where prompts are injected into multi-head self-attention (MSA) layers, used in existing prompt-based CL methods. (a) \textbf{Static gating} prepends prompts to a fixed, pre-defined subset of layers for all tasks and instances. (b) \textbf{Task-conditioned gating} learns a specific layer-wise topology per task but remains fixed for all instances within that task. (c) \textbf{Instance-conditioned gating (ours)} dynamically computes soft gate values for prompts at each layer for every input image via a learnable task-specific gating module. The varying intensity of each position illustrates our model's ability to adaptively weight layer-wise prompts based on the fine-grained visual context of each instance.}
    
    \label{fig:compare-prompt-position}
\end{figure}

A recent study analyzing the internal representations of ViT \citep{raghu2022visiontransformerslikeconvolutional} suggests that lower layers encode generic, low-level visual features associated with local attention patterns that are largely task-agnostic and transferable across tasks, while higher layers capture higher-level semantic concepts. This hierarchical distinction implies that uniformly injecting prompts across all layers, as done in OVOR \citep{huang2024ovor}, or in a fixed subset of early layers as in L2P \citep{wang2022learning} and \ta{CODA-Prompt} \citep{smith2023coda}, may be suboptimal. 
Indeed, DualPrompt empirically demonstrates that separating \textit{general} and \textit{expert} prompt locations improves performance. CODA-Prompt further improves prompt capacity through input-conditioned composition of decomposed prompt components, but follows a similar predefined prompt-insertion strategy. From a topology perspective, these methods still adopt a \emph{static prompting topology} or \emph{static gating} (see Figure \ref{fig:compare-prompt-position}a), where prompts are injected into every layer or a fixed subset of blocks. Such designs apply the same prompt activation pattern to all samples, leading to uniform computational and parameter usage regardless of task difficulty or input complexity.
RainbowPrompt \citep{hong2025rainbowprompt} partially relaxes this limitation by introducing task-conditioned probabilistic gates that learn where to insert prompts across layers for each task (Figure \ref{fig:compare-prompt-position}b). This task-level topology improves over manually predefined prompt locations and complements its diversity-enhanced prompt evolution mechanism. However, its gating mechanism remains \emph{task-conditioned}, the learned topology is shared across all inputs within a task, ignoring instance-level variations that may require adaptation at different depths. 

Motivated by this observation, we introduce an input-conditioned prompt gating mechanism (see Figure~\ref{fig:compare-prompt-position}c) that enables more precise layer-wise prompt activation for each individual input. This mechanism establishes an instance-specific topology that dynamically adjusts to the visual characteristics of each image. Inputs requiring fine-grained visual details may trigger prompts in the early layers with greater emphasis to refine low-level features, whereas images that align well with the frozen backbone may rely more on prompts in higher layers to facilitate label mapping. Furthermore, the gated prompts from previous tasks are dynamically aggregated with the current task prompt, enabling effective knowledge reuse across tasks through a dynamic knowledge fusion process. We hypothesize that this instance-conditioned prompting topology improves (1) the model's ability to capture fine-grained intra-task variations, and (2) long-term knowledge retention by selectively engaging only relevant historical prompt knowledge for each input. 
Besides improving instance-level adaptability, this design is also prompt-efficient: it avoids dense component-based composition and additional prompt-evolving transformation modules, relying instead on compact task-specific prompts and lightweight gating modules.

The main contributions of our work are as follows:
\begin{itemize}
\item We propose a novel prompt-based CL method, gated adaptive prompting (GAP-Prompt), which introduces input-conditioned gates to modulate layer-wise prompt activation for each instance. Moreover, by progressively integrating prompts from previous tasks with the current task prompt, GAP-Prompt forms an instance-specific prompting topology that enables adaptive knowledge reuse across tasks.

\item Our method achieves state-of-the-art (SoTA) results across multiple CL benchmarks, consistently improving accuracy while mitigating forgetting and maintaining stable performance under diverse task variations.
\end{itemize}

\section{Related works}
\label{sec:related works}

\subsection{Continual learning}
CL aims to incrementally acquire new knowledge while preserving previously learned information. The primary challenge in CL is catastrophic forgetting, where optimizing for new tasks leads to a severe decline in performance on earlier ones. To address this issue, traditional solutions are generally categorized into three main approaches: (1) \textit{regularization-based} methods mitigate forgetting by introducing penalty terms that protect the weights critical to prior tasks \citep{kirkpatrick2017overcoming, zenke2017continual, aljundi2018memory, benzing2022unifying, li2017learning, anh_wacv_2025, dangSAMixCalibratedAccurate2026}; (2) \textit{replay-based} methods maintain performance by either storing a subset of old data or using generative techniques to reconstruct past information \citep{rebuffi2017icarl, aljundi2019gradient, chaudhry2021using, buzzega2020darkexperiencegeneralcontinual}; (3) \textit{architecture-based} methods dynamically adjust the network structure or isolate specific parameter clusters to accommodate each new task \citep{mallya2018piggyback, serra2018overcoming, yoon_2018}. While these methods have made significant strides, the emergence of pre-trained models \citep{dosovitskiy2020image} trained on massive datasets has triggered a paradigm shift. Recent trends have moved away from resource-intensive retraining toward parameter-efficient approaches (e.g., LoRA~\citep{hu2021loralowrankadaptationlarge}, Prompt Tuning~\citep{lester2021powerscaleparameterefficientprompt}) and replay-free methods \citep{wang2022learning, wang2022dualprompt, smith2023coda}, harnessing powerful pre-trained generalization capabilities for more efficient and scalable learning.

% \citep{rusu_2016, yoon_2018, li_learn_tg_2019},

\subsection{Prompt-based continual learning}

For vision-based tasks, the prompting mechanism is typically implemented within the ViT \citep{dosovitskiy2020image} architecture, which relies on a hierarchical sequence of multi-head self-attention (MSA) \citep{vaswani2017attention} blocks. Prompting originally emerged as a method of using hand-crafted, task-specific instructions to elicit desired responses from pre-trained models. In the realm of CL, prompt-based strategies have gained significant traction by enabling the retention of task-specific knowledge without the need for a memory-intensive replay buffer. These methods \citep{wang2022learning, wang2022dualprompt, smith2023coda} treat prompts as keys to retrieve stored knowledge, effectively eliminating the requirement for constant model parameter updates. Within this domain, several pivotal frameworks have advanced the field. L2P \citep{wang2022learning} introduce a prompt-tuning mechanism that utilizes key-query matching to retrieve relevant task knowledge from a shared pool, while DualPrompt \citep{wang2022dualprompt} refines this by decoupling knowledge into task-invariant prompts for shared features and task-specific prompts to separately encode distinct task-related information. Coda-Prompt \citep{smith2023coda} proposes a decomposed attention-based prompting scheme to improve flexibility and scalability. 

Although the above methods prioritize the integration of multiple base prompts, they often overlook the representational diversity required for effective result unification. RainbowPrompt \citep{hong2025rainbowprompt} solves this by aggregating independent task prompts into a unified structure. However, while RainbowPrompt optimizes knowledge integration, its reliance on static statistics often leads to suboptimal triggering. Our approach mitigates this by leveraging the specific context of each input to ensure more precise and adaptive prompt activation.

\section{Methodology} 

\subsection{Preliminaries}
\subsubsection{Continual learning setting}
General CL consists of a sequence of $T$ tasks $\{1,...,T\}$. Let $\mathcal{D}_t$ as the dataset for $t$-th task with $N_t$ pairs: $\mathcal{D}_t = \{(\mathbf{x}_i, y_i)\}_{i=1}^{N_t}$. We focus on the class-incremental learning (class-IL) setting, where there is no overlap of classes between tasks, and task identity is unknown during inference.
% We focus on two popular settings in CL: class-incremental learning (Class-IL) and domain-incremental learning (Domain-IL). 
Besides, as in other prompt-based CL methods \citep{wang2022dualprompt, wang2022learning}, we adopt the rehearsal-free scenario, where no past samples are stored.

\subsubsection{Prompt-based method} 
Denote the pre-trained ViT model as $f_{\theta}$ with $L$ MSA layers, and the weights $\theta$ are frozen to prevent forgetting of base knowledge. With this standard ViT, the input to each transformer layer can be represented as:
\begin{equation}
    %  X = [x_{cls}; x_1; ...; x_m],
     \mathbf{X} = [\vx_{cls}; \mathcal{X}],
\end{equation}
where the input sequence is composed of a class token $\vx_{cls} \in \mathbb{R}^d$ and $m$ patch embeddings $\mathcal{X} \in \mathbb{R}^{d \times m}$. Prompt-based method augments this sequence with $n$ learnable prompt tokens $\vp \in \mathbb{R}^{d \times n}$. The reformulated input sequence is:
\begin{equation}
\label{eqn:vpt}
%   X = [x_{cls}; p_1; ...; p_n; x_1; ...; x_m].
  \mathbf{X} = [\vx_{cls}; \mathcal{X}; \vp].
\end{equation}

In addition, most prompt-based methods \citep{wang2022dualprompt, huang2024ovor} adopt \textit{prefix tuning} \citep{Li2021PrefixTuningOC}, where the prompts comprising $\vp^K \in \mathbb{R}^{n/2 \times d}$ and $\vp^V \in \mathbb{R}^{{n/2} \times d}$ are prepended to the key and value representations in the MSA layers of ViT. The \textit{prefix tuning} of the reformulated input sequence is defined as:
\begin{equation}
\label{eqn:prefix}
%   X = [x_{cls}; p_1; ...; p_n; x_1; ...; x_m].
%   X = [\vx_{cls}; \mathcal{X}; \vp].
  f_{prefix-tuning}(\mathbf{X}) =  \text{MSA}\left( \vx_{cls}, \mathbf{X}^Q, \begin{bmatrix} \vp^K \\ \mathbf{X}^K \end{bmatrix}, \begin{bmatrix} \vp^V \\ \mathbf{X}^V \end{bmatrix} \right),
\end{equation}
where $\mathbf{X}^Q, \mathbf{X}^K, \mathbf{X}^V$ are the query, key, and value matrices of the input embeddings. During training, only the prompt parameters $\vp^K$ and $\vp^V$ are updated, while all other model parameters remain frozen.

\subsubsection{Prompt position injection strategy}
Determining where prompts should be injected across MSA layers plays an important role in prompt-based CL. A recent approach inserts prompts uniformly across all transformer layers \citep{huang2024ovor}, increasing model capacity but also introducing additional computational overhead. Other methods restrict prompt injection to a fixed subset of early layers \citep{wang2022learning, wang2022dualprompt, smith2023coda}, motivated by the observation that lower layers capture more transferable visual representations. However, identifying suitable insertion layers is non-trivial and often relies on manual design choices or hyperparameter tuning. Moreover, such strategies result in static prompt insertion patterns that remain fixed across all inputs, ignoring instance-specific variations in visual complexity. 

Recently, RainbowPrompt \citep{hong2025rainbowprompt} addresses this challenge by introducing task-conditioned gating to learn prompt insertion layers for each task. Specifically, it learns a task-specific gate $\mathbf{G}^t = \{\mathbf{g}_l^t\}_{l=1}^{L}$, where $\mathbf{g}_l^t$ is a Bernoulli random variable indicating whether a prompt is inserted into layer $l$. To enable gradient-based optimization despite the discrete nature of $\mathbf{g}_l^t$, the method adopts the Gumbel-Softmax trick to obtain a differentiable relaxation: 
\begin{equation}
    \hat{g}_l^t(p) = 
    \frac{
    \exp\left(\log \delta_l^t(p) + Z_l^t(p)/\tau \right)
    }{
    \sum_{i \in \{0,1\}} 
    \exp\left(\log \delta_l^t(i) + Z_l^t(i)/\tau \right)
    }
    \label{eq:gumbel-softmax-rainbowprompt}
\end{equation}
where $p \in \{0,1\}$, $Z_l^t$ denotes the Gumbel noise, and $\tau$ is the temperature scaling parameter. During several initial training epochs, soft decisions sampled from the distribution $\delta_l^t = [\beta_l^t, 1-\beta_l^t]$ are used for optimization, where $\beta_l^t$ denotes the probability of inserting a prompt at layer $l$. After this early training phase, discrete gating decisions are sampled from the learned distribution and then fixed. Nevertheless, the learned topology is still shared across all inputs within the task, limiting the model's ability to adapt prompt activation to the specific characteristics of each instance.

\begin{figure}[t]
    \centering
    \includegraphics[width=\linewidth]{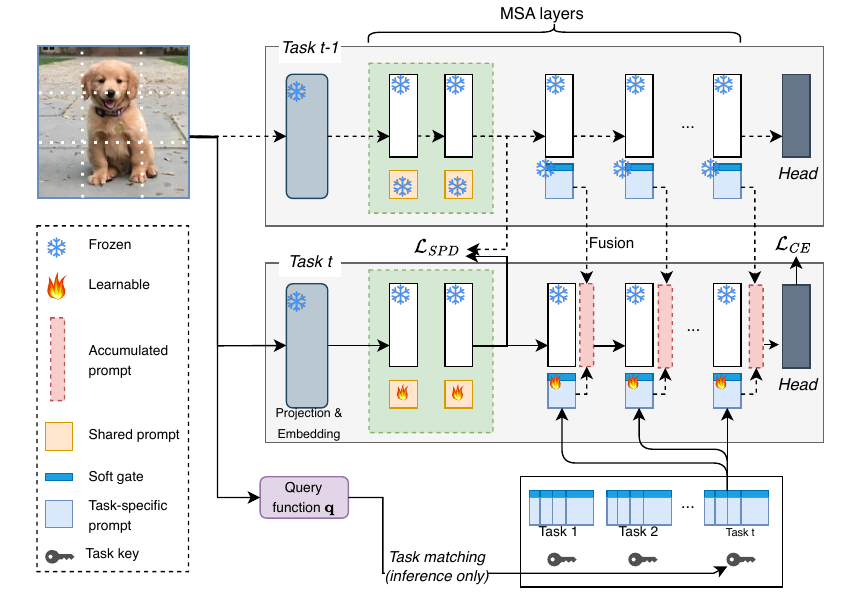}
    \caption{Overall architecture of GAP-Prompt. It partitions prompts into shared and task-specific components and comprises three main modules. Shared prompt distillation - SPD (green boxes and $\mathcal{L}_{SPD}$) stabilizes foundational knowledge in the shared prompt space. For task-specific prompts, instance-conditioned gating - ICG (blue bars Soft Gates) generates image-specific soft weights that drive dynamic knowledge fusion - DKF (pink boxes), which combines current and historical prompts into a unified representation. In inference, the query $\vq$ adaptively retrieves and aggregates prompt-gate pairs for each input.}
    \label{fig:gap-prompt-overall-architecture}
\end{figure}

\subsection{Overall architecture}
\label{subsec:overall-architecture}
% While we adopt a hierarchical approach from \citep{wang2022dualprompt} by partitioning prompts into shared and task-specific components, our framework introduces several critical advancements. 
The overall architecture of our method is illustrated in Figure~\ref{fig:gap-prompt-overall-architecture}. During training, each input image $\vx$ is first divided into patches and fed into the pre-trained ViT encoder. Following the hierarchical design of DualPrompt \citep{wang2022dualprompt}, we partition prompts into shared and task-specific expert components, denoted as $\vs_j^t$, $\vp_l^t$ for the current task $t$, respectively ($j$, $l$ denote layer indices). Following \citet{wang2022dualprompt}, shared prompts are injected into the first two layers and are learned across tasks, as these layers mainly capture general visual features that are broadly transferable \citep{raghu2022visiontransformerslikeconvolutional}. In contrast to DualPrompt, instead of fixing the next several layers to insert expert prompts, our method dynamically learns where to activate these prompts dynamically. To achieve this, we introduce an instance-conditioned gating (ICG) mechanism that generates soft activation weights for prompts at each layer. Similar to \citet{wang2022dualprompt}, a task-specific key $\vk^t$ is learned for each task $t$ to associate with its corresponding expert prompts. 
Moreover, previously learned shared and expert prompts are frozen. Unlike DualPrompt, where shared prompts are updated naively across tasks, we introduce a shared prompt distillation (SPD) loss to anchor the output representations of the first two layers from previous tasks and mitigate forgetting. In addition, for each input image $\vx$, the proposed ICG module computes dynamic soft gates that determine how prompts from different tasks contribute to the prompt representation of the current task. Together with frozen prompts from previous tasks, these gates enable dynamic knowledge fusion (DKF) mechanism, which combines the current learnable prompt with historical prompts. The resulting [CLS] token representation from the final transformer layer is fed into a classification head and optimized using the cross-entropy (CE) loss. \textit{During inference}, we first extract a query $\vq$ from the input image. A task-matching module then identifies the corresponding task-specific expert prompts. Conditioned on $\vq$, the model adaptively weights and aggregates these gated prompts to produce an input-specific prompt representation.

% NOTE: put current image to old prompt??

\subsection{Gated adaptive prompting mechanism}

\subsubsection{Instance-conditioned gating (ICG)}
\label{sec:icg}

Although the task-specific learnable gates in RainbowPrompt \citep{hong2025rainbowprompt} automatically learn, an optimized layer-wise topology for each task, they remain constrained by applying a fixed, static set of sampled gates to all instances within the task. This rigidity is suboptimal for handling high intra-task diversity and relies on stochastic sampling, which may permanently discard potentially useful prompts.
The proposed ICG addresses this limitation in a more effective and generative manner: instead of relying on a task-level sampled mask, it dynamically computes gating logits for each specific input image, which eliminates the rigidity of task-shared topological constraints. Concretely, instead of modeling prompt insertion as a discrete Bernoulli decision, we adopt a continuous gating formulation that directly represents the contribution strength of each expert prompt. To obtain differentiable gate values in $[0,1]$, we employ the Gumbel-sigmoid relaxation. Unlike the two-state softmax to sample a binary insertion decision, our formulation learns a continuous gate value for each layer, enabling multiple expert prompts to contribute simultaneously according to their relevance to the current input. Specifically, the gate $g_{l}^t$ for each instance $\mathbf{x}$ is defined as:

\begin{equation}
    g_{l}^t(\mathbf{x}) = \sigma(\frac{log(\pi_{l}^t(\mathbf{x})) + \epsilon}{\tau}),
    \label{eq:gate-value}
\end{equation}
% \begin{equation}
    
%     \label{query-x-q}
% \end{equation}
where $\sigma$ denotes the sigmoid function, $\epsilon$ is Gumbel noise sampled from a Gumbel(0, 1) distribution: using an uniform noise $u \sim uniform(0, 1)$, $\epsilon = -log(-log(u))$, and $\tau$ is a temperature parameter controlling the sharpness of the relaxation. Here, $\epsilon$ plays the same role as the Gumbel noise used in Eq.(~\ref{eq:gumbel-softmax-rainbowprompt}), but is simplified to a single noise term under the sigmoid-based relaxation. The gating parameter $\pi_l^t(\mathbf{x})$ is dynamically generated from the input representation: $\pi_l^t(\mathbf{x}) = h_{\phi}^{t}(\vq)$, where $\vq \in \mathbb{R}^d$ denotes the query of the input image $\vx$. $h_{\phi}^t$ denotes the network to learn the gates for task $t$, implemented as a simple linear layer. Besides, $\vq$ is extracted using the pre-trained ViT model $f_{\theta}$ as the feature vector of the [class] token: $\vq = f_{\theta}(\vx)[0]$.
We provide additional design details of $h_{\phi}^t$ in Appendix C.1.
% ~\ref{app:gating-module-architecture-design}. 
Together with the expert prompts $\vp_l^t$ at each layer $l$, the gating network $h_{\phi}^t$ is frozen after learning task $t$.

In addition, to control the sharpness of the relaxation, we employ a linear temperature annealing schedule. Specifically, the temperature $\tau$ is decreased over the training epochs,
\begin{equation}
    \tau = \tau_{start} - \frac{e-1}{E-1}(\tau_{start}-\tau_{end}),
    \label{eq:decreased-tau}
\end{equation}
where $e$ is the current epoch index (from 1) and $E$ is the total number of training epochs. 
Starting from a high temperature $\tau_{start}$ encourages exploratory soft gates, while gradually decreasing $\tau$ toward $\tau_{end}$ sharpens the relaxation and makes prompt activations more distinguishable without forcing them into strict 0s or 1s. 
This naturally matches our inference strategy: we keep the continuous gate values but suppress weak activations below the small threshold $\zeta$, allowing relevant prompts to remain active while filtering out irrelevant ones.

\subsubsection{Dynamic knowledge fusion (DKF)}
While RainbowPrompt \citep{hong2025rainbowprompt} shows that higher representation diversity can improve accuracy and reduce forgetting, its prompt integration mechanism produces a unified prompt that remains fixed during inference. This static representation limits the model's ability to adapt prompt utilization to individual inputs. To overcome this limitation, we propose dynamic knowledge fusion (DKF), which dynamically aggregates historical and current prompts using gates produced by the ICG mechanism. Conditioned on the visual characteristics of each input instance, these gates determine the contribution of each prompt, enabling input-adaptive knowledge fusion. Furthermore, unlike RainbowPrompt, which introduces an additional attention-based transformation module with extra parameters and increases training complexity, DKF simply computes the previously learned instance-conditioned gates and only learn the gates for the current task. This design keeps the fusion mechanism simple while enabling knowledge retention across tasks with instance-level adaptability.

Specifically, after learning each task, the corresponding prompt is frozen. Let $\mathcal{P}_{l}^{past} = \{\vp_l^1, \vp_l^2,...,\vp_l^{t-1} \}$ denotes the set of old frozen expert prompts from the $t-1$ previously learned tasks at the $l$-th MSA layer.  
% Here, each prompt element in the set $\mathcal{P}_{l}^{past}$ is frozen - not learnable during current task (see Section~\ref{subsec:overall-architecture}). 
Considering new task $t$ at $l$-th layer, we have new learnable prompt $\vp_l^t$, with corresponding learnable gate $g^t_l(\vx)$. For a given input image $\vx$ with query feature $\vq$, we dynamically aggregate all old prompts in $\mathcal{P}_{l}^{past}$ and current prompt $\vp_l^t$ based on their corresponding ICG activation weight $g_l^i(\vx)$. Note that old gate $\{g_l^1,\ldots,g_l^{t-1}\}$ are computed with given input $\vx$ via Eq.~(\ref{eq:gate-value}). The fused prompt $\mathbf{a}_l^t$ used for current task is formulated as a normalized weighted sum:
\begin{equation}
    \mathbf{a}_l^t(\vx) = \frac{\sum_{i=1}^{t-1}{g_l^i(\vx)}{\vp_l^i} + g_l^t(\vx)\vp_l^t}{\sum_{i=1}^{t-1}{g_l^i(\vx)} + g_l^t(\vx) + \eta}
    \label{eq:a-l-t}
\end{equation}
% (we used $1e-8$)
where $\eta$ is a small constant added for numerical stability to prevent division by zero. The aggregated prompt $\va_l^t$ is then computed on the fly for the current task $t$, and is not stored. Instead, $\vp_l^t$ is stored for each task, with $\va_l^t$ being recalculated during both training and inference. Separating the $i = t$ term highlights that $\{\vp_l^t,g_l^t\}$ are learnable for current task $t$ while $\{\vp_l^i,g_l^i\}_{i<t}$ are frozen.

This formulation in Eq.~(\ref{eq:a-l-t}) provides two key benefits. First, the instance-conditioned weights ${g}^t_l(\vx)$ create a {\textit{soft ensemble}}, allowing the model to selectively draw knowledge from relevant past prompts instead of relying on a rigid task-shared representation. Second, normalizing the accumulated prompt by the total activated gate weight acts as a {\textit{distribution stabilizer}}, keeping the injected prompt magnitude bounded and consistent with the representation scale of the pre-trained ViT backbone.

\textbf{Comparison with prior prompt integration methods.}
GAP-Prompt is related to CODA-Prompt~\citep{smith2023coda} and RainbowPrompt~\citep{hong2025rainbowprompt}, but differs in the role and granularity of prompt adaptation. 
CODA-Prompt uses input-conditioned coefficients for dense prompt-component composition, whereas GAP-Prompt uses sigmoid-bounded gates as activation strengths that control where and how strongly current and historical task prompts are injected into MSA layers. 
RainbowPrompt performs task-level prompt evolution with task-specific layer gates, while GAP-Prompt fuses historical prompts on the fly using instance-conditioned layer-wise gates. 
Thus, GAP-Prompt shifts prompt integration from dense component composition or task-level evolution to instance-level sparse routing and normalized knowledge fusion. 
Further formulation, parameter, and efficiency analyses are provided in Appendix C.2 and E. 

\subsubsection{Shared prompt distillation (SPD)}
% Shared Prompt Knowledge Distillation
Since we employ shared prompts in the first two layers as in \citet{wang2022dualprompt}, forgetting in these prompts is inevitable. To mitigate forgetting, we propose to adopt an additional SPD loss in the output of these two layers. Specifically, after learning the first task, the SPD loss is defined as:

\begin{equation}
    \mathcal{L}_{SPD} = 1 - \gamma(\mathbf{z}^{t}, \mathbf{z}^{t-1}),
    \label{eq:spd}
\end{equation}

where $\gamma(\cdot,\cdot)$ denotes the cosine similarity between two feature representations. Here, $\mathbf{z}^{t-1}$ and $\mathbf{z}^{t}$ represent the features of the first two MSA layers produced by the pre-trained ViT with previous shared prompts $\vs_j^{t-1}$ and the current shared prompts $\vs_j^{t}$ ($j \in \{1,2\}$), respectively. Minimizing this loss encourages the current shared prompts to preserve the feature representations learned by previous tasks.

\subsubsection{Overall training objective}
To learn the key $\vk^{t}$ for the current task $t$, we use the standard key matching loss as in \citet{wang2022dualprompt}:
    \begin{equation}
        \mathcal{L}_{match}(\vx, \vk^t) = \gamma (\vq, \vk^t), \qquad \vx \in \mathcal{D}_t
        \label{eq:original-key-matching-loss}
    \end{equation}
where $\gamma(\cdot, \cdot)$, similar to Eq.~(\ref{eq:spd}), denotes cosine similarity, and $\vq$ is the query feature extracted from the input $\vx$. The final learning objective of our method is:
    \begin{equation}
        \mathcal{L}_{final} = \mathcal{L}_{CE} + \lambda_{match} \mathcal{L}_{match} + \lambda_{SPD} \mathcal{L}_{SPD}
    \end{equation}

where $\mathcal{L}_{CE}$ denotes the standard cross-entropy classification loss, $\lambda_{match}$ and $\lambda_{SPD}$ are hyperparameters that control the contributions of the key matching loss and the SPD loss, respectively.

\subsubsection{Inference process}
\label{subsubsec:inference-process}
During inference, we employ a small deterministic activation threshold $\zeta$ (empirically set to 0.1) to discretize the gate logits, suppressing negligible activation values. This filtering step benefits the proposed DKF mechanism by removing weak and noisy prompt contributions during fusion. As a result, only prompts with sufficient relevance to the current input participate in the aggregation process. Consequently, ICG produces more reliable gating decisions, and the fusion process becomes more stable and robust during inference.

Specifically, the inference procedure consists of the following steps.

\textit{(1) Task querying.} Given an input image $\vx$, the corresponding query $\vq$ is extracted
then 
%VN: no need repeatly saying we follow others. Following \cite{dualprompt}, 
$\vq$ is compared with all task keys $\{\vk^1, \vk^2,\ldots,\vk^t\}$ to identify the closest key with index $\hat{t}$:

\begin{equation}
    % \hat{t} = argmax_{i}(\gamma (\vq, \vk^i)),
    \hat{t} = \operatorname*{arg\,max}_{i \in \{1,\ldots,t\}} \gamma(\vq, \vk^i)
    \label{eq:inference-task-matching}
\end{equation}

\textit{(2) Prompt retrieval.} Given the predicted task index $\hat{t}$, the prompt used for each layer $l$ is aggregated as:

\begin{equation}
    \va_l^{\hat{t}}(\vx) = \frac{\sum_{i=1}^{\hat{t}}{g_l^i(\vx)}{\vp_l^i}}{\sum_{i=1}^{\hat{t}}{g_l^i(\vx)} + \eta}
\end{equation}

Here, the gate $g_l^i$ for each task $i \in \{1,\ldots,\hat{t}\}$ is calculated from the input image $\vx$ using Eq.~(\ref{eq:gate-value}) with threshold $\zeta$ and without Gumbel noise $\epsilon$, and is associated with the corresponding frozen prompt $\vp_l^i$. Finally, the fused prompt $\va_l^{\hat{t}}$, together with the shared prompts $\vs$ at the early layers, is used for inference.

\section{Experiments}
\subsection{Datasets and implementation details}

\noindent \textbf{Datasets.} To evaluate the performance of GAP-Prompt, we conduct comprehensive evaluations on image classification in the class-incremental learning settings across three primary benchmarks: ImageNet-R \citep{hendrycks2021many}, CIFAR-100 \citep{krizhevsky2009learning}, and CUB-200 \citep{welinder2010caltech}. ImageNet-R is selected for its inherent difficulty, containing 200 classes with diverse artistic styles and high intra-class variance. CIFAR-100 (100 classes) is a standard benchmark for prompt-based CL to evaluate generalization. Finally, we utilize the fine-grained CUB-200 dataset of 200 bird species to test the model's ability to distinguish subtle inter-class differences. Across all experiments, each dataset is partitioned into 10 tasks with disjoint classes: 20 classes/task for ImageNet-R, 10 classes/task for CIFAR-100, and 20 classes/task for CUB-200.

\noindent \textbf{Evaluation metrics.} 
To evaluate performance, we use two standard metrics: Average Accuracy and Forgetting \citep{chaudhry_2018}. Average Accuracy measures the model's performance across all tasks after completing the training sequence, while Forgetting quantifies the performance drop on previously learned tasks. Let $R_{t,i}$ denote the accuracy on task $i$ after training on task $t$. The Average Accuracy $\mathcal{A}_T$ after the final task $T$ is defined as: $\mathcal{A}_T = \frac{1}{T}\sum_{k=1}^{T} R_{T, k}$.
% \begin{equation}
%     \mathcal{A}_T = \frac{1}{T}\sum_{k=1}^{T} R_{T, k}
% \end{equation}
Similarly, the Forgetting metric $\mathcal{F}$ is defined as the mean performance drop from each task's initial accuracy to its final state: $\mathcal{F} = \frac{1}{T-1} \sum_{i=1}^{T-1}(R_{i,i}-R_{T,i})$.

\noindent \textbf{Compared methods.} We evaluate the performance of GAP-Prompt against a comprehensive suite of methods, including non-prompt-based methods such as LAE \citep{gao2023lae}, InfLoRA \citep{liang2024inflora}, and EASE \citep{zhou2024ease}, as well as established prompt-based architectures: L2P \citep{wang2022learning}, DualPrompt \citep{wang2022dualprompt}, Coda-Prompt \citep{smith2023coda}, ESN \citep{wang2023isolation}, EvoPrompt~\citep{kurniawan2024evolving}, OVOR~\citep{huang2024ovor}, CPrompt~\citep{gao2024consistent}, and RainbowPrompt~\citep{hong2025rainbowprompt}.

\noindent \textbf{Implementation details.} 
All experiments are executed on a dual NVIDIA V100 GPUs.
Following \citet{hong2025rainbowprompt, wang2022learning}, we primarily use the supervised pre-trained backbones on ViT-B/16 on ImageNet-1K \citep{russakovsky2015imagenet} for CIFAR-100 and ImageNet-R and ImageNet-21K \citep{ridnik2021imagenet} for CUB-200. Besides, we also utilize self-supervised pretrained backbones, including iBOT-1K~\citep{zhou2022ibotimagebertpretraining} and DINO-1K~\citep{caron2021emergingpropertiesselfsupervisedvision}. 
GAP-Prompt is optimized using the AdamW optimizer \citep{loshchilov2017decoupled} with a cosine learning rate decay schedule to stabilize training. For training configurations, we employ a batch size of 64 for ImageNet-R, while a larger batch size of 128 is utilized for both CIFAR-100 and CUB-200. In addition, we re-implement RainbowPrompt using the official public code with the same experimental settings.

\subsection{Experimental results} 
\label{sec:result}

\subsubsection{Main results}
Results on three benchmarks 
% under the 10-task Class-IL setting
are summarized in Table~\ref{table:main_results}.
\begin{itemize}
    \item \textbf{CIFAR-100:} On CIFAR-100, {GAP-Prompt} achieves a new SoTA average accuracy of {89.24\%}, surpassing the previous best prompt-based method, EvoPrompt \citep{kurniawan2024evolving}, by a margin of {+1.27\%}. Furthermore, our method achieves the lowest forgetting rate of {3.03\%}, demonstrating that the synergy between SPD and DKF effectively anchors historical knowledge. Compared to RainbowPrompt \citep{hong2025rainbowprompt} (87.22\%), which utilizes static task-specific gating, our instance-conditioned approach yields a significant improvement, highlighting the benefit of image-specific adaptation.
    
    \item \textbf{ImageNet-R:} For the larger and more diverse ImageNet-R dataset, {GAP-Prompt} maintains its lead with an average accuracy of {78.72\%}, slightly outperforming competitive baselines like CPrompt \citep{gao2024consistent} (77.14\%) and EvoPrompt \citep{kurniawan2024evolving} (76.83\%). Our method effectively manages the representation shift in large-scale scenarios, reporting the lowest forgetting rate on this benchmark ({3.12\%}). This consistency across scales validates the robustness of our gated adaptive prompting mechanism. 
   
    \item \textbf{CUB-200:} The most striking results are observed on CUB-200, a challenging fine-grained classification task. {GAP-Prompt} reaches an average accuracy of {87.29\%}, which is remarkably close to the joint training upper bound of {88.00\%}. Our method outperforms the previous best-reported prompt-based method, OVOR-Deep \citep{huang2024ovor}, by a substantial {+9.18\%}, and represents a staggering {+16.86\%} improvement over RainbowPrompt \citep{hong2025rainbowprompt}. The drastic reduction in forgetting ({3.68\%}) compared to other methods suggests that our method is particularly adept at capturing the subtle, instance-specific visual nuances required to distinguish between similar sub-categories over time.
\end{itemize}

Across all three datasets, \text{GAP-Prompt} consistently achieves the highest accuracy and lowest forgetting. These results show that ICG, normalized DKF, and SPD jointly enable effective instance-wise prompt aggregation, allowing the model to exploit accumulated prompt knowledge, preserve old knowledge, and remain adaptable to new tasks.

\begin{table*}[!t]
\begin{center}

\caption{Performance comparison with other methods on CIFAR-100, ImageNet-R, and CUB-200 under the 10-task class-IL setting. Best results are in \textbf{bold}.} 
\setlength{\tabcolsep}{1mm}
\resizebox{\linewidth}{!}{
\begin{tabular}{lcccccc}
\toprule 
\multirow{2}{*}{\textbf{Method}} & \multicolumn{2}{c}{\textbf{CIFAR-100}}
 &  \multicolumn{2}{c}{\textbf{ImageNet-R}}&  \multicolumn{2}{c}{\textbf{CUB-200}}\\ 
\cmidrule(lr){2-3} \cmidrule(lr){4-5} \cmidrule(lr){6-7}  % Adds horizontal lines above each pair
& \multicolumn{1}{c}{Avg Acc($\uparrow$)} & \multicolumn{1}{c}{Forgetting($\downarrow$)}   & \multicolumn{1}{c}{Avg Acc($\uparrow$)} & \multicolumn{1}{c}{Forgetting($\downarrow$)}& \multicolumn{1}{c}{Avg Acc($\uparrow$)} & \multicolumn{1}{c}{Forgetting($\downarrow$)} \\

\midrule
Joint training & 91.38 & -- & 82.06 & -- & 88.00 & -- \\
\midrule

 LAE~\citep{gao2023lae} & 86.09 \tsb{\( \pm \)0.55} & 6.22 \tsb{\( \pm \)0.17} & 73.88 \tsb{\( \pm \)0.43} & 5.08 \tsb{\( \pm \)0.33} & - & - \\
 InfLoRA~\citep{liang2024inflora} & 86.91 \tsb{\( \pm \)0.34} & 4.22 \tsb{\( \pm \)0.37} & 76.13 \tsb{\( \pm \)0.44} & 5.22 \tsb{\( \pm \)0.48} & - & - \\
 EASE~\citep{zhou2024ease} & 87.28 \tsb{\( \pm \)0.71} & 4.81 \tsb{\( \pm \)0.42} & 75.82 \tsb{\( \pm \)0.54} & 5.83 \tsb{\( \pm \)0.37} & - & - \\
 
\midrule

L2P~\citep{wang2022learning}  &83.86 \tsb{\( \pm \)0.28} & 7.35 \tsb{\( \pm \)0.38}& 61.57 \tsb{\( \pm \)0.66} & 9.73 \tsb{\( \pm \)0.47}  & 71.22 \tsb{\( \pm \)0.54}  & 10.68 \tsb{\( \pm \)0.61} \\

DualPrompt~\citep{wang2022dualprompt} & 86.51 \tsb{\( \pm \)0.33} & 5.16 \tsb{\( \pm \)0.09} &  68.13 \tsb{\( \pm \)0.49}  & 4.68 \tsb{\( \pm \)0.20}  &71.55 \tsb{\( \pm \)0.73}  & 10.21 \tsb{\( \pm \)0.43} \\

CODA-Prompt~\citep{smith2023coda} & 86.92 \tsb{\( \pm \)0.77}& 5.10 \tsb{\( \pm \)0.43}&   73.21 \tsb{\( \pm \)0.28}  & 5.69 \tsb{\( \pm \)0.28} & 73.25 \tsb{\( \pm \)0.71} & 11.02 \tsb{\( \pm \)0.88} \\

ESN~\citep{wang2023isolation} & 86.42 \tsb{\( \pm \)0.80} & 6.08 \tsb{\( \pm \)0.48} & 75.11 \tsb{\( \pm \)0.36}  & 5.68 \tsb{\( \pm \)0.77}  & 71.20 \tsb{\( \pm \)0.54} & 9.82 \tsb{\( \pm \)0.43} \\

EvoPrompt~\citep{kurniawan2024evolving} & 87.97 \tsb{\( \pm \)0.30} & 3.12 \tsb{\( \pm \)0.50} &  76.83 \tsb{\( \pm \)0.08}    & 3.34 \tsb{\( \pm \)0.07} & - & -  \\

OVOR-Deep~\citep{huang2024ovor} & 85.99 \tsb{\( \pm \)0.89} & 6.42 \tsb{\( \pm \)2.03}&  76.11 \tsb{\( \pm \)0.21}    & 7.16 \tsb{\( \pm \)0.34} & 78.11 \tsb{\( \pm \)0.47}   & 7.95 \tsb{\( \pm \)0.77} \\

CPrompt~\citep{gao2024consistent} & 87.82 \tsb{\( \pm \)0.21} & 5.06 \tsb{\( \pm \)0.50}& 77.14 \tsb{\( \pm \)0.11}   & 5.97 \tsb{\( \pm \)0.68}  & 77.09  \tsb{\( \pm \)0.64}& 10.27 \tsb{\( \pm \)0.54} \\

% VQ-Prompt & \text{88.73 \tsb{\( \pm \)0.27}} & - & \text{78.71 \tsb{\( \pm \) 0.22} } & - & \text{86.72 \tsb{\( \pm \)0.94} } & - \\

RainbowPrompt~\citep{hong2025rainbowprompt} & \text{87.22 \tsb{\( \pm \)0.53}} & 4.81 \tsb{\( \pm \)0.41} & \text{76.65 \tsb{\( \pm \) 0.56} } & 4.57 \tsb{\( \pm \) 0.33} & \text{70.43 \tsb{\( \pm \)1.23} } & 10.53 \tsb{\( \pm \)0.91} \\

\midrule

GAP-Prompt (Ours) & \textbf{89.24 \tsb{\( \pm \)0.28}} & \textbf{3.03 \tsb{\( \pm \)0.44}} & \textbf{78.72 \tsb{\( \pm \)0.35} } & \textbf{3.12 \tsb{\( \pm \)0.54}} & \textbf{87.29 \tsb{\( \pm \)0.45} } & \textbf{3.68 \tsb{\( \pm \)0.27}} \\

\bottomrule
\end{tabular}
}

\vspace{-2ex}
\label{table:main_results}
\end{center}
\end{table*}

\subsubsection{Performance with self-supervised pre-trained models}
\label{app:per-with-ss-pretrained}
\begin{table*}[!t]
    \begin{center}
        \caption{Average accuracy on CIFAR-100 and ImageNet-R with self-supervised pre-trained models iBOT-1K and DINO-1K. Best results are in \textbf{bold}.} 
        \label{tab:results_pretrained_v2}
        \setlength{\tabcolsep}{6mm} 
        \resizebox{\linewidth}{!}{
        \begin{tabular}{llcc}
            \toprule 
            \textbf{Pre-trained model} & \textbf{Method} & \textbf{CIFAR-100} & \textbf{ImageNet-R} \\
            \cmidrule(lr){1-1} \cmidrule(lr){2-2} \cmidrule(lr){3-3} \cmidrule(lr){4-4} 
            \multirow{5}{*}{iBOT-1K} 
            & L2P  & $75.57 \pm 0.41$ & $60.97 \pm 0.26$ \\
            & DualPrompt  & $76.63 \pm 0.05$ & $61.51 \pm 1.05$ \\
            & CODA-Prompt  & $79.11 \pm 1.02$ & $66.56 \pm 0.68$ \\
            & RainBowPrompt  & $76.20 \pm 0.34$ & $65.82 \pm 0.52$ \\
            & \textbf{GAP-Prompt (Ours)} & $\mathbf{79.90 \pm 0.57}$ & $\mathbf{69.29 \pm 0.32}$ \\
            
            \midrule
            
            \multirow{5}{*}{DINO-1K} 
            & L2P  & $70.65 \pm 0.57$ & $57.40 \pm 0.23$ \\
            & DualPrompt  & $74.90 \pm 0.21$ & $58.57 \pm 0.45$ \\
            & CODA-Prompt  & $77.50 \pm 0.64$ & $63.15 \pm 0.39$ \\
            & RainBowPrompt  & $74.77 \pm 0.54$ & $62.04 \pm 0.43$ \\
            & \textbf{GAP-Prompt (Ours)} & $\mathbf{77.61 \pm 0.38}$ & $\mathbf{66.46 \pm 0.45}$ \\
            
            \bottomrule
        \end{tabular}
        }
    \end{center}
\end{table*}

Tab.~\ref{tab:results_pretrained_v2} presents a performance comparison between GAP-Prompt and current SoTA methods on the CIFAR-100 and ImageNet-R benchmarks, utilizing pre-trained backbones including iBOT-1K~\citep{zhou2022ibotimagebertpretraining} and DINO-1K~\citep{caron2021emergingpropertiesselfsupervisedvision}. 
The experimental results demonstrate that GAP-Prompt achieves superior and consistent improvements across all testing scenarios. 

Specifically, with iBOT-1K, GAP-Prompt reaches $79.90\%$ on CIFAR-100 and $69.29\%$ on ImageNet-R, outperforming CODA-Prompt by $+0.79\%$ and $+2.73\%$, respectively. 
Similarly, with DINO-1K, our method remains competitive on CIFAR-100 and achieves a clear $+3.31\%$ gain over CODA-Prompt on ImageNet-R ($66.46\%$ vs. $63.15\%$). 
Notably, the consistent gap over RainbowPrompt across both backbones further suggests that instance-conditioned gating is more effective than task-level gating in adapting prompt usage to the visual context of each input. 
Overall, these results demonstrate the robustness and generalization ability of GAP-Prompt under different pre-training paradigms.

% \subsubsection{Efficiency comparison with task-specific adaptive prompting}

\subsubsection{Robustness comparison with task-specific adaptive prompting}

\begin{figure}
    \centering
    \includegraphics[width=1.0\linewidth]{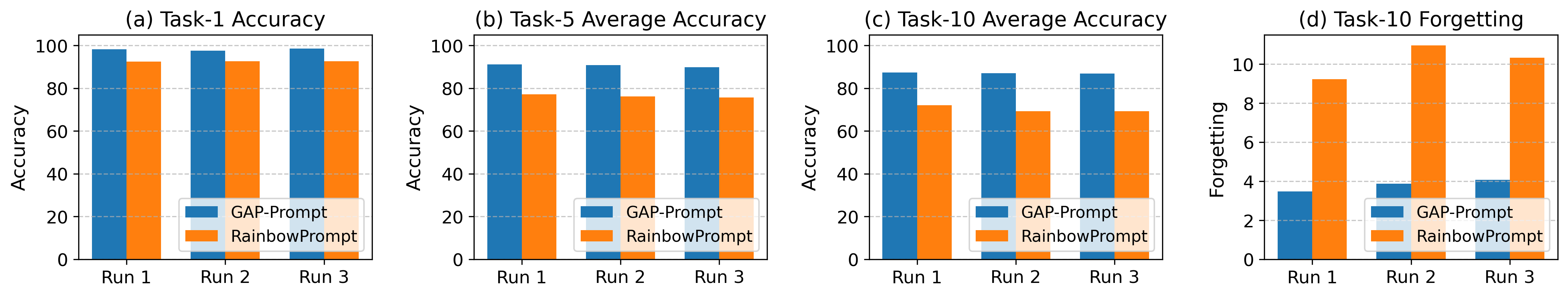}
    \caption{Comparison between our method (GAP-Prompt) and task-based adaptive prompting (RainbowPrompt \citep{hong2025rainbowprompt}) on the dataset CUB-200, 10 tasks, CIL setting. (a) Accuracy after learning the first task; (b) Average accuracy (AA) after learning the fifth task; (c) AA after learning the final task; (d) Forgetting of the model after the final task. }
    \label{fig:comparison-with-static-gating}
\end{figure}

To validate our hypothesis that ICG improves over task-based adaptive prompting~\citep{hong2025rainbowprompt}, we compare GAP-Prompt with RainbowPrompt under varying class splits on CUB-200. The results, shown in Figure~\ref{fig:comparison-with-static-gating}, highlight the impact of these variations on task performance.

\begin{itemize}
    
    \item \textbf{After completing the first task:} (Figure~\ref{fig:comparison-with-static-gating}a), where no forgetting occurs, our method consistently outperforms RainbowPrompt. 
    This shows the benefit of ICG alone: it selects prompts based on each input instead of using a task-shared prompt subset. 
    The stable accuracy across runs further indicates that instance-conditioned layer-wise activation provides more consistent learning than task-based adaptive prompting.
    
    \item \textbf{After completing the fifth task:} (Figure~\ref{fig:comparison-with-static-gating}b), our method shows the same trend: it consistently outperforms RainbowPrompt and remains stable across runs. At this stage, in addition to ICG, both DKF and SPD are also involved, further demonstrating the superiority of our approach over RainbowPrompt.
    
    \item \textbf{After completing the tenth task:} (Figure~\ref{fig:comparison-with-static-gating}c, Figure~\ref{fig:comparison-with-static-gating}d), our method still maintains a superior result compared to RainbowPrompt. Furthermore, our method remains stable across runs, while RainbowPrompt's results fluctuate, indicating its instability when the diversity of tasks changes. Figure~\ref{fig:comparison-with-static-gating}d further highlights this effect, showing that RainbowPrompt exhibits a much higher forgetting rate than GAP-Prompt and varies significantly across runs when the class set in each task changes, in contrast to GAP-Prompt's low forgetting rate and minimal variation.
\end{itemize}

\subsection{Ablation study}

\subsubsection{Component-wise ablation}

To evaluate the individual contribution of each proposed component in our framework, we conduct a comprehensive ablation study on the CIFAR-100 dataset. Starting from a baseline configuration that excludes our core components (ICG, DKF, and SPD), we systematically evaluate the impact of our method by incrementally integrating these components. The results, summarized in Table~\ref{tab:ablation_with_time}, highlight the synergistic effect of these components in enhancing accuracy and mitigating forgetting.

\begin{table}[htbp]
\centering
\caption{Component-wise ablation of GAP-Prompt  on the CIFAR-100 dataset}
\label{tab:ablation_with_time}
\begin{tabular*}{\textwidth}{l@{\extracolsep{\fill}}cc}
\toprule
\textbf{Training Strategy} & \textbf{Avg Acc} ($\uparrow$) & \textbf{Forgetting} ($\downarrow$) \\
\midrule
Baseline                                & 84.78 $\pm$ 0.80 & 4.56 $\pm$ 0.73 \\
+ Instance-Conditioned Gating (ICG)     & 87.82 $\pm$ 0.37 & 4.05 $\pm$ 0.22 \\
+ Dynamic Knowledge Fusion (DKF)        & 88.94 $\pm$ 0.46 & 3.73 $\pm$ 0.31 \\
+ Shared Prompt Distillation (SPD)      & \textbf{89.24 $\pm$ 0.28} & \textbf{3.03 $\pm$ 0.44} \\
\bottomrule
\end{tabular*}
\end{table}

\paragraph{Instance-conditioned gating (ICG).}
Integration of ICG marks the most significant performance jump, improving average accuracy from 84.78\% to 87.82\% (a gain of +3.04\%) and reducing forgetting from 4.56\% to 4.05\%. This substantial improvement validates our hypothesis that a static, task-level gating topology is insufficient for handling high intra-task diversity. 
By allowing each instance to dynamically navigate the layer-wise prompt space via ICG, the model captures fine-grained visual features more effectively.

\paragraph{Dynamic knowledge fusion (DKF).}
By adding DKF upon the ICG-enabled backbone, the average accuracy further increases to 88.94\%, while forgetting drops to 3.73\%. DKF leverages the soft weights produced by ICG to perform context-aware aggregation, allowing the model to selectively reuse relevant knowledge from prompts of previous tasks based on the current input. The normalized weighted sum also stabilizes the representation distribution, keeping the fused features aligned with the pre-trained ViT.

\paragraph{Shared prompt distillation (SPD).}
The inclusion of SPD provides the final refinement, achieving the highest average accuracy of 89.24\% and the lowest forgetting rate of 3.03\%. While the early layers of a ViT are known to capture generic, task-agnostic patterns, they remain susceptible to overwriting during sequential learning. SPD effectively anchors the shared prompts in these initial layers by minimizing the cosine similarity between current and previous features of the first two MSA layers. Although the accuracy gain is modest (+0.30\%), the noticeable reduction in forgetting demonstrates its critical role in preserving previous knowledge across tasks.

Consequently, the combination of these three modules allows our framework to achieve a superior balance: ICG provides the necessary \textit{plasticity} to adapt to diverse instances, while DKF and SPD provide the \textit{stability} required to preserve knowledge over long-term sequences. Together, they outperform the baseline by a total of $4.46\%$ in accuracy while simultaneously reducing forgetting by $1.53\%$.

\subsubsection{Prompt position granularity}
To evaluate the effectiveness of \textit{instance-conditioned gating (ICG)} in GAP-Prompt against the \textit{task-based adaptive prompting (TAP)} mechanism in RainbowPrompt~\citep{hong2025rainbowprompt}, we use a simplified fixed-position prompting baseline without ICG, DKF, and SPD. We then integrate either TAP or ICG into this baseline for comparison. As reported in Tab.~\ref{tab:appendix_extra_ablation}, ICG significantly improves the fixed-position baseline and outperforms TAP by $1.25\%$ in average accuracy while reducing forgetting by $1.63\%$. These results demonstrate the effectiveness of ICG in enabling instance-level dynamic activation of layer-wise prompts, rather than relying on fixed prompt positions (as in the baseline) or task-based adaptive prompting (as in TAP). This supports our claim that ICG provides finer-grained control over layer-wise prompt activation, rather than merely performing weighted prompt summation. When combined with DKF and SPD in the full GAP-Prompt framework, performance further improves. This indicates that instance-level routing, controlled historical knowledge fusion, and shared-prompt stabilization are complementary, enabling GAP-Prompt to clearly outperform RainbowPrompt.

\begin{table}[htbp]
    \centering
    \caption{Comparison between ICG and TAP on CIFAR-100. \textbf{Bold} highlights the best results.}
    \label{tab:appendix_extra_ablation}
    \begin{tabular*}{\textwidth}{l@{\extracolsep{\fill}}cc}
        \toprule
        \textbf{Method} & \textbf{Avg Acc} ($\uparrow$) & \textbf{Forgetting} ($\downarrow$) \\
        \midrule
        Baseline (\textit{fixed prompting})                                & 84.78 $\pm$ 0.80 & 4.56 $\pm$ 0.73 \\
        Baseline + \textit{Task-based adaptive prompting (TAP)} & 86.62 $\pm$ 0.30 & 5.68 $\pm$ 0.26 \\
        Baseline + \textit{Instance-conditioned gating (ICG)}     & 87.82 $\pm$ 0.37 & 4.05 $\pm$ 0.22 \\
        \midrule
        RainbowPrompt~\citep{hong2025rainbowprompt}      & {87.22 $\pm$ 0.53} & {4.81 $\pm$ 0.41} \\
        \ta{GAP}-Prompt      & \textbf{89.24 $\pm$ 0.28} & \textbf{3.03 $\pm$ 0.44} \\
        \bottomrule
    \end{tabular*}
\end{table}

\section{Conclusion}

In this paper, we introduce GAP-Prompt, an efficient and robust framework for prompt-based continual learning. By transitioning from rigid task-level topologies to an instance-conditioned gating mechanism, GAP-Prompt effectively captures fine-grained intra-task variances often neglected by static prompting methods (both fixed prompt injections and task-specific adaptive prompting). Our framework combines adaptive plasticity (learning new knowledge) through ICG, historical knowledge integration via dynamic knowledge fusion, and foundational stability (preserving previously learned knowledge) through shared prompt distillation. Extensive experiments on CIFAR-100, ImageNet-R, and CUB-200 demonstrate the superiority of GAP-Prompt over existing state-of-the-art methods in both accuracy and forgetting mitigation. In particular, the strong performance on fine-grained bird species classification highlights the framework's ability to adapt to complex representation spaces. We believe GAP-Prompt provides a scalable and effective paradigm for lifelong learning in dynamic, real-world environments. In future work, we plan to explore more challenging settings, such as online continual learning and few-shot class-incremental learning.

\bibliography{main}
\bibliographystyle{tmlr}

\appendix

\include{appendix/appendix}
% \section{Appendix}

\end{document}

%% file: math_commands.tex
\usepackage{amsmath,amsfonts,bm}

\def\eqref#1{equation~\ref{#1}}
\def\1{\bm{1}}

\def\va{{\bm{a}}}

\def\vk{{\bm{k}}}

\def\vp{{\bm{p}}}
\def\vq{{\bm{q}}}

\def\vs{{\bm{s}}}

\def\vx{{\bm{x}}}

\DeclareMathAlphabet{\mathsfit}{\encodingdefault}{\sfdefault}{m}{sl}
\SetMathAlphabet{\mathsfit}{bold}{\encodingdefault}{\sfdefault}{bx}{n}

%% file: appendix/appendix.tex
\section{Hyperparameter selection} 
The hyperparameters used in \ta{GAP-Prompt} are described in Tab.~\ref{tab:search-space-hyperparameters}. 
For selecting optimal hyperparameters, we utilize a grid search method. The final chosen values for all hyperparameters are shown in Tab.~\ref{tab:chosen-samix-hyperparameters}. In addition to these hyperparameters, in \ta{GAP-Prompt}, we inject shared prompts into the first two layers following the hierarchical design in \citet{wang2022dualprompt}, and we inject task-specific expert prompts into the next eight layers with instance-conditioned gates $g_l^t$ to dynamically activate these prompts. We do not inject any prompts into the last two layers, as \citet{raghu2022visiontransformerslikeconvolutional} reveals that, in these highest layers, the network strictly focuses on updating the [CLS] token for the final classification readout, while spatial tokens (representing image patches or regions) are mostly bypassed via skip connections. Injecting task-specific expert prompts at this stage disrupts the dedicated feature aggregation process, leading to negligible or even negative benefits while increasing computational overhead. When re-implementing RainbowPrompt~\citep{hong2025rainbowprompt}, to ensure a fair comparison, we utilize the configurations and hyperparameters specified in the original paper.

\begin{table}[h]
    \caption{Hyperparameters used in \ta{GAP-Prompt}}
    \centering
    \setlength{\tabcolsep}{4mm}
    \resizebox*{!}{0.34\linewidth}{
        % \small
        % \input{AnonymousSubmission/LaTeX/tables/table_search_space_hyperparameters}
        \begin{tabular}{cl}
            \toprule
            \textbf{Hyperparameter} & \textbf{Description} \\
            \midrule
            $E$  & Number of training epochs \\
            $\tilde{\eta}$ & Learning rate \\
            $\lambda_{match}$ & Coefficient of $\mathcal{L}_{match}$ in the final loss \\
            $\lambda_{SPD}$ & Coefficient of $\mathcal{L}_{SPD}$ in the final loss \\
            $bsz$ & Batch size \\
            $N_{\vp}$ & Task-specific expert prompt length \\
            $N_{\vs}$ & Shared prompt length \\
            $\zeta$ & Threshold for removing noise in gates $g_l^t$ during inference \\
            $\tau_{start}$, $\tau_{end}$ & Initial and terminate value of temperature used to \\
            & compute gate $g_l^t$ \\
            \bottomrule
        \end{tabular}
    }
    \label{tab:search-space-hyperparameters}
\end{table}

\begin{table}[h]
    \caption{Selected hyperparameters in \ta{GAP-Prompt}}
    \centering
    \setlength{\tabcolsep}{8mm}
    \resizebox*{!}{0.25\linewidth}{
        \small
        \begin{tabular}{cc}
            \toprule
            \textbf{Dataset} & \textbf{Hyperparameter} \\
            \midrule
            All             & $N_{\vp}: 20$, $N_{\vs}: 6$, $\lambda_{match}: 1.0$, \\
                            & $\tau_{start}: 5.0$, $\tau_{end}: 0.1$, $\zeta: 0.1$ \\
            \midrule
            CIFAR-100       & $\tilde{\eta}: 0.005$, $E: 20$, $bsz: 128$, $\lambda_{SPD}: 0.1$ \\
            \midrule
            ImageNet-R      & $\tilde{\eta}: 0.003$, $E: 50$, $bsz: 64$, $\lambda_{SPD}: 0.01$ \\
            \midrule
            CUB-200         & $\tilde{\eta}: 0.005$, $E: 20$, $bsz: 128$, $\lambda_{SPD}: 0.1$ \\
            \bottomrule
        \end{tabular}
    }
    \label{tab:chosen-samix-hyperparameters}
\end{table}

\section{Algorithm overview}

% The training algorithm and the inference procedure of our method are summarized in Algo.~\ref{algo:training} and Algo.~\ref{algo:inference}, respectively. All equation references in the algorithms refer to equations in the main paper.

Our training and inference procedures are summarized in Algo.~\ref{algo:training} and Algo.~\ref{algo:inference}, respectively. 
All equation references in the algorithms refer to the main paper.

\begin{minipage}{0.9\textwidth}
    \begin{algorithm}[H]
    \caption{\emph{GAP-Prompt training}}
    \label{algo:training}
    
    \KwIn{Training sets $\{\mathcal{D}_t\}_{t=1}^{T}$, frozen pre-trained ViT $f_\theta$, hyperparameters $\lambda_{match}, \lambda_{SPD}$}
    
    \BlankLine
    Initialize shared prompts $\{\vs_j\}$
    
    \BlankLine
    \For{$t = 1,\ldots,T$}{
        Initialize task-specific expert prompts $\{\vp_l^t\}$, task key $\vk^t$, linear gating module $h_\phi^t$, and classification head $\tilde{h}_\psi^t$
        
        \If{$t > 1$}{
            Store and freeze the old shared prompts $\{\vs_j^{t-1}\} \gets \{\vs_j\}$
        }
        
        \BlankLine
        \For{$(\vx, y) \in \mathcal{D}_t$}{
            
            % Extract query $\vq$ following Eq.~(6)
            Extract query $\vq \gets f_{\theta}(\vx)[0]$

            \tcp{Instance-conditioned gating (ICG)}
            % \LeftComment{1}{\text{Instance-conditioned gating (ICG)}}
            Generate instance-conditioned gates $g_l^t(\vx)$ using linear module $h_\phi^t$ as Eq. (5)
            \BlankLine
            
            \tcp{Dynamic knowledge fusion (DKF)}
            Perform DKF to get $\va_l^t(\vx)$ following Eq. (7)
            
            Forward $\vx$ through ViT $f_{\theta}$ using $\vs_j$ and $\va_l^t(\vx)$ as per Eq. (3)
            \BlankLine
            
            Compute $\mathcal{L}_{match}$ following Eq. (9) and $\mathcal{L}_{CE}$;
            \BlankLine
            
            \tcp{Shared prompt distillation (SPD)}
            \If{$t > 1$}{
                Compute $\mathcal{L}_{SPD}$ with $\{\vs_j^{t-1}\}$ following Eq. (8)
            }
            
            Update $\{\vs_j\}$, $\{\vp_l^t\}$, $\vk^t$, $h_\phi^t$, $\tilde{h}_\psi^t$ by minimizing $\mathcal{L}_{final}$ in Eq. (10)
        }
        \BlankLine
        \BlankLine
        Freeze $\{\vp_l^t\}$, $\vk^t$, $h_\phi^t$
    }
    \KwOut{Updated shared prompts $\{\vs_j\}$, expert prompts $\{\vp_l^t\}$, task key $\vk^t$, linear module $h_\phi^t$, and classification heads $\tilde{h}_\psi^t$}
    \end{algorithm}
\end{minipage}

% \setcounter{AlgoLine}{0}

% --- Algorithm 2: GAP-Prompt Inference ---
\begin{minipage}{0.9\textwidth}
    \begin{algorithm}[H]
        \setcounter{AlgoLine}{0} 
        \caption{\emph{GAP-Prompt inference after learning task $t$}}
        \label{algo:inference}
        
        \KwIn{Input image $\vx$, frozen pre-trained ViT $f_\theta$, learned shared prompts $\{\vs_j\}$, task-specific expert prompts $\{\vp_l^t\}$, linear gating module $h_{\phi}^t$, and classification head $\tilde{h}_\psi^t$, threshold $\zeta$}
        
        \BlankLine
        % Extract query $\vq$ from input $\vx$ following Eq. (6)\;
        Extract query $\vq \gets f_{\theta}(\vx)[0]$ \;
        
        Perform \textit{task querying} to identify task index $\hat{t}$ following Eq. (11)\;
        
        \BlankLine
        \For{$i = 1,\ldots,\hat{t}$}{
            Compute instance-conditioned gates $g_l^i(\vx)$ using linear module $h_\phi^i$ following Eq. (5) and filter with threshold $\zeta$\;
        }
        
        \BlankLine
        Perform \textit{prompt retrieval} and fusion $\va_l^{\hat{t}}(\vx)$ following Eq. (12)\;
        
        Forward $\vx$ through ViT $f_\theta$ with fused prompts $\va_l^{\hat{t}}(\vx)$ to get final features\;
        
        Compute $logits$ across all learned classes using classification head $\tilde{h}_\psi^t$ \;
        
        Compute predicted label $\hat{y} = \text{argmax}(logits)$\;
        
        \BlankLine
        \KwOut{Predicted label $\hat{y}$}
    \end{algorithm}
\end{minipage}

% Algorithm 2: GAP-Prompt Inference
% --- Algorithm 2: GAP-Prompt Inference Pipeline ---
% \begin{algorithm}
% \caption{Inference Pipeline for GAP-Prompt}
% \label{alg:gap_inference}
% \begin{algorithmic}[1]
% \STATE \textbf{Input:} Test sample $\mathbf{x}$, Trained ViT $f_\theta$, General prompts $\mathbf{G}$, Expert Pool $\{\mathbf{P}, \mathbf{K}\}$, Gating functions $G$.
% \STATE \textbf{Output:} Predicted class label $\hat{y}$.
% \STATE Extract global query feature $\mathbf{q}$ following eq. (7).
% \STATE Find predicted task index $\hat{t}$ via key matching following eq. (10).
% \FOR{each expert layer $l \in \mathcal{L}_{expert}$}
%     \STATE Compute hard gates $\{g_l^1, \dots, g_l^{\hat{t}}\}$ following eq. (3).
%     \STATE Aggregate prompts to obtain fused representation $\mathbf{a}_l^{\hat{t}}(\mathbf{x})$ following eq. (11).
%     \STATE Prepend $\mathbf{a}_l^{\hat{t}}$ and shared prompts $\mathbf{s}$ to MSA layers.
% \ENDFOR
% \STATE $\hat{y} = \arg\max (\text{ClassifierHead}(f_\theta(\mathbf{x})[\text{CLS}]))$.
% \STATE \textbf{Return} $\hat{y}$.
% \end{algorithmic}
% \end{algorithm}

\section{Design choices and stability analysis}
\label{app:design_choice_stability}

This section provides additional analyses for the main design choices of GAP-Prompt. We first compare the proposed lightweight linear gate with a bottleneck MLP for the gating network in Sec.~\ref{app:gating-module-architecture-design}. 
% We then discuss Gumbel-Sigmoid gating, temperature annealing, and inference thresholding for sparse prompt routing in Sec.~\ref{app:why-gumbel-sigmoid}. Finally, we analyze normalized Dynamic knowledge fusion (DKF) and its role in stabilizing fused prompt magnitudes across tasks in Sec.~\ref{app:normalized_fusion}.
We then provide a formulation-level comparison to clarify how GAP-Prompt differs from prior prompt integration methods in Sec.~\ref{app:comparison_prior_prompt_integration}. 
Specifically, we analyze the role of ICG compared with dense attention-based component weighting in CODA-Prompt~\citep{smith2023coda} and task-level gating in RainbowPrompt~\citep{hong2025rainbowprompt}, and then discuss why Gumbel-Sigmoid is suitable for ICG. 
Finally, we analyze why DKF uses normalized fusion for instance-conditioned historical prompt integration.

\subsection{Gating module architecture design}
\label{app:gating-module-architecture-design}

% To learn the gating network in GAP-Prompt, intuition and standard practices in parameter-efficient fine-tuning (e.g., Adapters \cite{houlsby2019parameter}) often suggest that a bottleneck Multi-Layer Perceptron (MLP) with non-linear activations yields a more expressive and noise-resilient mechanism. In general representation learning, this compression-expansion structure effectively filters out redundant information and captures complex non-linear mappings. To verify this, we replace our linear layer with a bottleneck MLP (consists of following layers: Down-projection linear, normalization, GELU activation function, Up-projection linear). The results on CUB-200 dataset are shown in Table 1. Interestingly, we observed a performance drop. We attribute this to two factors: first, the bottleneck heavily compresses the rich input-aware features, destroying fine-grained context necessary for instance-level routing. Second, the normalization and non-linearities dampen the unconstrained logit variance, leading to overly smoothed, ambiguous gate activations rather than the sharp, deterministic routing required for optimal prompt selection. Therefore, keeping the gate strictly linear ensures both high-fidelity matching and computational efficiency.

To learn the gating network in GAP-Prompt, intuition and standard practices in parameter-efficient fine-tuning, such as Adapters \citep{houlsby2019parameternlp}, might suggest that a bottleneck multi-layer perceptron (MLP) with non-linear activations provides a more expressive and noise-resilient mechanism. Such compression–expansion structures are widely used in representation learning to filter redundant information capture complex non-linear mappings.

\begin{table*}[ht]
\begin{center}

\caption{Comparison of gating module architectures in GAP-Prompt (Linear and MLP) on CIFAR-100, ImageNet-R, and CUB-200 under the 10-task class-IL setting. The best results are highlighted in \textbf{bold}.} 
\setlength{\tabcolsep}{1mm}
\resizebox{\linewidth}{!}{
\begin{tabular}{lcccccc}
\toprule 
\multirow{2}{*}{\textbf{Gating module}} & \multicolumn{2}{c}{\textbf{CIFAR-100}}
 &  \multicolumn{2}{c}{\textbf{ImageNet-R}}&  \multicolumn{2}{c}{\textbf{CUB-200}}\\ 
\cmidrule(lr){2-3} \cmidrule(lr){4-5} \cmidrule(lr){6-7}  % Adds horizontal lines above each pair
& \multicolumn{1}{c}{Avg Acc($\uparrow$)} & \multicolumn{1}{c}{Forgetting($\downarrow$)}   & \multicolumn{1}{c}{Avg Acc($\uparrow$)} & \multicolumn{1}{c}{Forgetting($\downarrow$)}& \multicolumn{1}{c}{Avg Acc($\uparrow$)} & \multicolumn{1}{c}{Forgetting($\downarrow$)} \\
\midrule

MLP & \text{88.28 \tsb{\( \pm \)0.20}} & \text{3.69 \tsb{\( \pm \)0.18}} & \text{76.27 \tsb{\( \pm \) 0.36} } & \text{4.35 \tsb{\( \pm \) 0.33}} & \text{86.32 \tsb{\( \pm \)0.36} } & \text{4.10 \tsb{\( \pm \)0.11}} \\

Linear & \textbf{89.24 \tsb{\( \pm \)0.28}} & \textbf{3.03 \tsb{\( \pm \)0.44}} & \textbf{78.72 \tsb{\( \pm \)0.35} } & \textbf{3.12 \tsb{\( \pm \)0.54}} & \textbf{87.29 \tsb{\( \pm \)0.45} } & \textbf{3.68 \tsb{\( \pm \)0.27}} \\

\bottomrule
\end{tabular}
}

% \vspace{-2ex}
\label{table:linear-mlp-comparison}
\end{center}
\end{table*}

To verify this design, we replace the linear gating layer with a bottleneck MLP consisting of a down-projection linear layer, normalization, GELU activation, and an up-projection linear layer. The results on CIFAR-100, ImageNet-R, and CUB-200 are reported in Tab.~\ref{table:linear-mlp-comparison}. Interestingly, this modification leads to a performance drop compared to the linear gating layer. Specifically, the average accuracy decreases by 0.96\%, 2.45\%, and 0.97\% on CIFAR-100, ImageNet-R, and CUB-200, respectively, accompanied by increased forgetting. We attribute this degradation to two main factors. First, the bottleneck compresses the rich input-aware features, potentially discarding fine-grained information necessary for instance-level routing. Second, the normalization and non-linearities reduce the variance of the gating logits, resulting in overly smooth and ambiguous gate activations rather than the sharp routing signals required for effective prompt selection. 
Therefore, we retain a linear gating mechanism, which preserves high-fidelity feature matching while remaining computationally efficient.

\subsection{Why GAP-Prompt differs from prior prompt integration methods}
\label{app:comparison_prior_prompt_integration}
GAP-Prompt is closely related to CODA-Prompt~\citep{smith2023coda} and RainbowPrompt~\citep{hong2025rainbowprompt}, since all three methods aim to improve prompt integration in CL. 
However, they differ in two key aspects: (1) the weighting or activation mechanism for prompts, and (2) the way historical prompt knowledge is integrated. 
We first compare ICG with prior weighting and gating mechanisms, then analyze the normalized DKF formulation for historical prompt fusion.

\subsubsection{How ICG differs from prior weighting/gating mechanisms}
\label{app:icg_comparison}
Instance-conditioned gating (ICG) is designed to produce input-conditioned, layer-wise prompt activation strengths rather than dense attention coefficients or task-level insertion decisions. We clarify this design choice by comparing ICG with CODA-Prompt~\citep{smith2023coda} and RainbowPrompt~\citep{hong2025rainbowprompt} below.

\paragraph{Dense attention-based prompt weighting in CODA-Prompt.}
CODA-Prompt~\citep{smith2023coda} decomposes prompts into a set of prompt components and uses input-conditioned attention coefficients to compose the final prompt. For clarity and consistency with our notation, we denote $M$ as the number of new components introduced per task. After learning task $t$, the available component dictionary therefore contains $tM$ components, and the prompt at layer $l$ is formed as:

\begin{equation}
    \mathbf{p}^{\mathrm{CODA}}_l(\mathbf{x})
    =
    \sum_{m=1}^{tM}
    \alpha_{l,m}(\mathbf{x}) \mathbf{P}_{l,m},
    \label{eq:final-prompt-coda}
\end{equation}
where $\mathbf{P}_{l,m}$ is the $m$-th prompt component at layer $l$, and $\alpha_{l,m}(\mathbf{x})$ is the input-conditioned component weight. Each prompt component $\mathbf{P}_{l,m}$ is associated with a corresponding component key $\mathbf{K}_{l,m}$ and a learnable attention vector $\mathbf{A}_{l,m}$. When a new task arrives, CODA-Prompt expands the component dictionary by adding $M$ new triplets $\{(\mathbf{P}_{l,m}, \mathbf{K}_{l,m}, \mathbf{A}_{l,m})\}$, while previously learned components and their corresponding keys and attention vectors are frozen. This expansion strategy allows new capacity to be added without changing the weight computation of previously learned components. Here, the query derived from the input image $\mathbf{x}$ is denoted as $\mathbf{q}(\mathbf{x})$ instead of simply $\mathbf{q}$ to emphasize its input dependence. CODA-Prompt first computes an attended query by applying the component-specific attention vector $\mathbf{A}_{l,m}$ to the input query $\mathbf{q}(\mathbf{x})$, and then compares this attended query with the corresponding component key $\mathbf{K}_{l,m}$:
\begin{equation}
    \alpha_{l,m}(\mathbf{x})
    =
    \gamma
    \left(
    \mathbf{q}(\mathbf{x}) \odot \mathbf{A}_{l,m},
    \mathbf{K}_{l,m}
    \right),
\end{equation}
where $\gamma(\cdot,\cdot)$ denotes cosine similarity and $\odot$ denotes element-wise multiplication. Here, $\mathbf{A}_{l,m}$ is a learnable component-specific feature-selection vector rather than an input-conditioned module. It modulates the query representation before cosine matching with $\mathbf{K}_{l,m}$. 

Therefore, the role of $\alpha_{l,m}(\mathbf{x})$ in CODA-Prompt is prompt-component composition: it determines how decomposed components are assembled into the final prompt content. 
This differs from ICG, where the input-dependent value is not used to compose prompt components, but to decide the activation strength of a task-specific prompt at a particular layer.

% where $\mathbf{A}_{l,m}$ denotes the learnable attention vector associated with the $m$-th component and $\odot$ denotes element-wise modulation. Under this formulation, the input-conditioned weights act as \emph{composition coefficients}: they determine how prompt components are combined to form the final prompt content.

\paragraph{Task-based Gumbel-Softmax gating in RainbowPrompt.} 
RainbowPrompt~\citep{hong2025rainbowprompt} also studies prompt insertion across layers, but its gate is task-conditioned rather than instance-conditioned. Specifically, a task-specific Bernoulli gate is defined as:

\begin{equation}
    \mathbf{G}^t = \{\mathbf{g}_l^t\}_{l=1}^{L},
\end{equation}
where $g_l^t$ indicates whether the RainbowPrompt of task $t$ should be inserted into layer $l$. To optimize this discrete decision, RainbowPrompt uses a two-state Gumbel-Softmax relaxation:
\begin{equation}
    \hat{g}_l^t(p)
    =
    \frac{
    \exp\left((\log \delta_l^t(p) + Z_l^t(p))/\tau\right)
    }{
    \sum_{i\in\{0,1\}}
    \exp\left((\log \delta_l^t(i) + Z_l^t(i))/\tau\right)
    },
    \quad p\in\{0,1\},
\end{equation}
where $\delta_l^t=[\beta_l^t,1-\beta_l^t]$ parameterizes the probability of inserting or not inserting the prompt at layer $l$. After the early training phase, discrete gates are sampled and then fixed for that task. Therefore, all inputs belonging to the same task share the same layer-wise prompt insertion pattern.

\paragraph{Instance-conditioned gating (ICG) in GAP-Prompt.}
Unlike CODA-Prompt, which produces input-dependent coefficients to compose prompt contents from a component dictionary, and unlike RainbowPrompt, which learns task-level insertion gates shared by all samples of a task, ICG predicts an activation strength for each input and each layer via Gumbel-Sigmoid relaxation:

\begin{equation}
    g_l^t(\mathbf{x})
    =
    \sigma
    \left(
    \frac{
    \log(\pi_l^t(\mathbf{x})) + \epsilon
    }{\tau}
    \right),
\end{equation}
where $\epsilon$ is Gumbel noise, $\tau$ is the temperature, and $\sigma(\cdot)$ is the sigmoid function. 
The gating parameter is generated from the input query:
\begin{equation}
    \pi_l^t(\mathbf{x}) = h_\phi^t(\mathbf{q}),
\end{equation}
where $h_\phi^t$ is the task-specific gating module and $\mathbf{q}$ is the query feature extracted from the frozen ViT.

Since the output is constrained to $[0,1]$, $g_l^t(\mathbf{x})$ can be interpreted as the activation strength of the task-specific prompt $\mathbf{p}_l^t$ at layer $l$ for input $\mathbf{x}$. This design differs from RainbowPrompt in both conditioning and interpretation: two samples from the same task can produce different gate values, and the gate represents a continuous activation strength rather than a sampled binary insertion decision. 
Moreover, these gates are later used by DKF to control the fusion of current and historical prompts, instead of only deciding whether a task-level prompt is inserted into a layer.
Thus, CODA-Prompt asks which prompt components should be composed, RainbowPrompt asks whether a task-level prompt should be inserted into a layer, whereas ICG asks which layer-wise task prompt should be activated for this particular input and to what extent. 
This changes the role of the weight from dense prompt-component composition or task-level insertion to instance-conditioned sparse prompt routing.

% ICG differs from the design of gating in RainbowPrompt in both conditioning and interpretation. First, ICG is conditioned on the input query. Thus, two samples from the same task can produce different gate values. Second, ICG produces a continuous activation strength rather than a sampled binary insertion decision. This is important because different images may require different degrees of prompt intervention at different representation depths. Third, the gate is later used by DKF to control the fusion of current and historical prompts, rather than only deciding whether a task-level prompt is inserted into a layer.

\paragraph{Why Gumbel-Sigmoid for ICG?}
The goal of ICG is to learn discrete-like prompt activation decisions while keeping the gating module differentiable during training. 

 \textit{A standard sigmoid gate},
\begin{equation}
    g_l^t(\mathbf{x}) = \sigma(\rho_l^t(\mathbf{x})),
\end{equation}
where $\rho_l^t(\mathbf{x})$ denotes the gating logit predicted from the input query, 
% We use $\rho_l^t(\mathbf{x})$ here to distinguish this standard sigmoid baseline from the Gumbel-Sigmoid formulation, where $\pi_l^t(\mathbf{x})$ is used inside the log-transformed relaxed Bernoulli parameter.
can produce soft activation values, but may remain overly smooth, with limited exploration and no explicit encouragement toward sharp routing decisions.

\textit{A softmax-based attention mechanism}, as another possible alternative to ICG, assigns dense relative weights over the available task prompts:

% on the other hand, produces dense relative weights over candidates, forcing all prompts to compete within a normalized distribution. 

\begin{equation}
    w_l^i(\mathbf{x})
    =
    \frac{
    \exp(\gamma(\mathbf{q},\mathbf{k}^i_l))
    }{
    \sum_{j=1}^{t}
    \exp(\gamma(\mathbf{q},\mathbf{k}^j_l))
    },
    \quad
    \mathbf{a}_l(\mathbf{x})
    =
    \sum_{i=1}^{t}
    w_l^i(\mathbf{x}) \mathbf{p}_l^i.
\end{equation}

Here, this formulation would require maintaining an additional key $\mathbf{k}^i_l$ for each task prompt $\mathbf{p}_l^i$. Although it resembles attention-style dense prompt weighting, it is not equivalent to CODA-Prompt, where the weights are computed over decomposed prompt components with component-specific keys and attention vectors. 
More importantly, this dense formulation forces all candidate prompts to compete within a normalized distribution and allows every prompt to contribute to each input, even when some historical prompts are irrelevant. This behavior is less suitable for class-incremental learning, where irrelevant prompt reuse can introduce interference.

Based on the Gumbel-Softmax relaxation~\citep{jang2017categorical, maddison2017concrete}, Gumbel-Sigmoid better matches the intended behavior of ICG because it treats each prompt activation as an independent routing decision rather than a dense normalized competition among all candidates. 
The Gumbel perturbation encourages exploration of different layer-wise activation patterns during training, while the sigmoid relaxation keeps the gate value in $[0,1]$ and allows gradient-based optimization. 
As the temperature is annealed, the gates become sharper and more separable, making them naturally compatible with inference-time thresholding. 
Thus, Gumbel-Sigmoid supports the sparse routing objective of ICG: it learns soft gates during training, suppresses irrelevant prompts through thresholding, and provides routing signals for DKF.

\subsubsection{Why normalized dynamic knowledge fusion (DKF)?}
\label{app:why_normalized_dkf}
While ICG determines how strongly each layer-wise task prompt should be activated for a given input, DKF determines how current and historical task prompts are integrated. 
Here, we compare DKF with the prompt integration mechanisms in CODA-Prompt and RainbowPrompt, and then discuss why normalized fusion is used.

\paragraph{Dense component integration in CODA-Prompt.}
CODA-Prompt integrates prompt knowledge by densely composing decomposed prompt components as in Eq.~(\ref{eq:final-prompt-coda}).
Historical knowledge is reused because previously learned components remain in the component dictionary and can still contribute to future prompts through their input-conditioned coefficients. 
However, this integration is performed at the component level. 
The weights $\alpha_{l,m}(\mathbf{x})$ determine how decomposed prompt components are assembled into the final prompt, rather than deciding which task-specific historical prompts should be activated at each layer for each input.

\paragraph{Task-based prompt evolution in RainbowPrompt.}
RainbowPrompt integrates historical prompts through a prompt-evolving mechanism. 
Let $\{\mathbf{p}_l^1,\ldots,\mathbf{p}_l^t\}$ denote accumulated task prompts up to task $t$ at layer $l$. 
RainbowPrompt transforms and aligns these accumulated prompts during training:
\begin{equation}
    \tilde{\mathbf{p}}_l^i
    =
    \mathcal{A}_l^t(\mathbf{p}_l^i),
    \quad \{i=1,\ldots,t\},
\end{equation}
where $\mathcal{A}_l^t(\cdot)$ denotes the task-conditioned transformation and alignment operation used for prompt evolution. 
The evolved prompts are then integrated into a unified task-level RainbowPrompt:
\begin{equation}
    \mathbf{p}^{\mathrm{rainbow}}_{l,t}
    =
    \frac{1}{t}
    \sum_{i=1}^{t}
    \tilde{\mathbf{p}}_l^i.
\end{equation}
The resulting evolved prompts are then used at inference. Therefore, RainbowPrompt performs historical prompt integration at the task level: the evolved prompt is shared by samples from the same task and remains fixed during inference.

\paragraph{Instance-conditioned historical prompt fusion in GAP-Prompt.}
GAP-Prompt addresses a complementary problem: historical prompt relevance may vary across inputs, even within the same task. 
After learning task $i$, the prompt $\mathbf{p}_l^i$ and its gating module $h_\phi^i$ are frozen. 
For an input $\mathbf{x}$ at task $t$, Dynamic Knowledge Fusion (DKF) computes gates $g_l^i(\mathbf{x})$ for current and historical prompts and fuses them on the fly:
\begin{equation}
    \mathbf{a}_l^t(\mathbf{x})
    =
    \frac{
    \sum_{i=1}^{t}
    g_l^i(\mathbf{x})\mathbf{p}_l^i
    }{
    \sum_{i=1}^{t}
    g_l^i(\mathbf{x}) + \eta
    }.
\end{equation}
Unlike CODA-Prompt, the weights in DKF are not component-composition coefficients. 
They are activation strengths over task-specific prompts. 
Unlike RainbowPrompt, the fused prompt is not a stored task-level representation, but is recomputed for every input. 
Therefore, DKF performs instance-conditioned historical prompt fusion rather than dense component integration or task-level prompt evolution.

\paragraph{Why normalization?}
The denominator in DKF prevents the magnitude of the fused prompt from growing with the number of learned tasks or the number of active gates. 
Without normalization, an unnormalized gated sum,
\begin{equation}
    \bar{\mathbf{a}}_l^t(\mathbf{x})
    =
    \sum_{i=1}^{t}
    g_l^i(\mathbf{x})\mathbf{p}_l^i
\end{equation}
can increase in norm as more prompts are accumulated. 
This may shift the input distribution seen by the frozen ViT backbone, because prompt tokens are injected into the key and value streams of the MSA layers. 
Normalized DKF instead keeps the injected prompt scale controlled:
\begin{equation}
    \|\mathbf{a}_l^t(\mathbf{x})\|_2
    \leq
    \frac{
    \sum_{i=1}^{t}
    g_l^i(\mathbf{x})\|\mathbf{p}_l^i\|_2
    }{
    \sum_{i=1}^{t}
    g_l^i(\mathbf{x})+\eta
    }.
\end{equation}
When the prompt norms are bounded, the fused prompt norm is also controlled. 
This is important because the ViT backbone is frozen, and the learned prompts must remain compatible with the representation scale of the pre-trained model.

% =========================================================================
% ================ Convergence and stability of DKF ================
\paragraph{Convergence and stability of DKF.}
We further analyze the convergence behavior and optimization stability of Dynamic Knowledge Fusion (DKF) using training loss curves. 
Since DKF recomputes the fused prompt from current and historical prompts for each input, one potential concern is that the fusion process may introduce unstable optimization dynamics. 
We therefore visualize the training loss curves at intermediate and final stages, including task 5, task 10, and the global training loss across datasets.

\begin{figure}[t]
\centering
\includegraphics[width=0.95\linewidth]{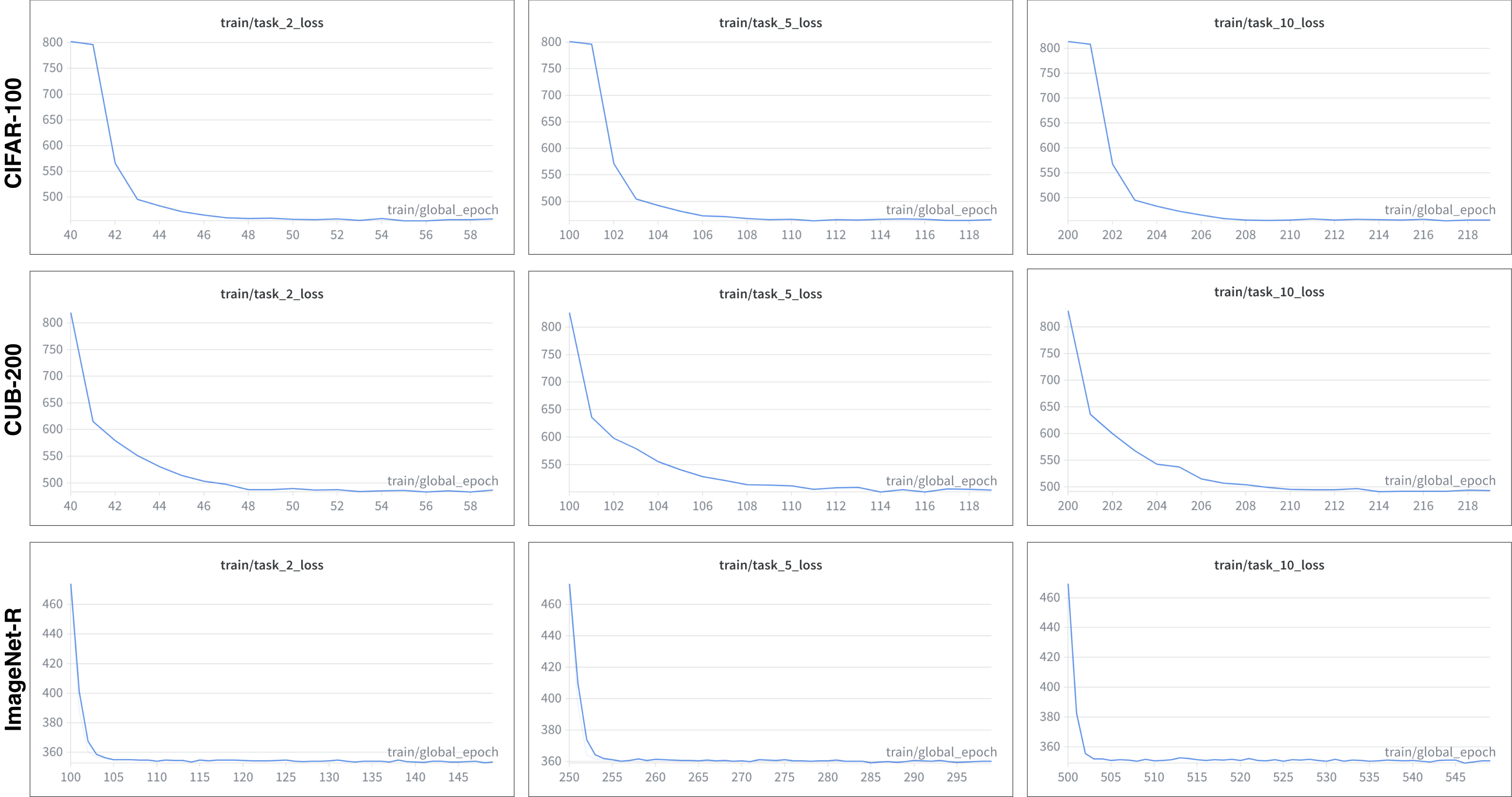}
\caption{
Convergence and optimization stability of DKF. 
We visualize the training loss curves during task 2, task 5, and task 10 training on different datasets. 
The curves decrease smoothly without severe oscillations, suggesting that input-specific dynamic fusion does not destabilize optimization.
}
\label{fig:dkf_loss_curves}
\end{figure}

As shown in Fig.~\ref{fig:dkf_loss_curves}, the loss decreases smoothly without severe oscillations. 
This suggests that DKF does not destabilize optimization, despite recomputing input-specific fused prompts during training. 
Together with the normalized formulation discussed above, this supports the stability of DKF from both an optimization perspective and a scale-control perspective.

\input{appendix/c3_param_efficiency_considerations}
\subsubsection{Summary of differences}
Table~\ref{tab:formulation_comparison} summarizes the key differences among CODA-Prompt~\citep{smith2023coda}, RainbowPrompt~\citep{hong2025rainbowprompt}, and GAP-Prompt. CODA-Prompt uses input-conditioned weights for dense prompt component composition. RainbowPrompt integrates historical prompts through task-level prompt evolution and uses task-level layer gates. In contrast, GAP-Prompt uses instance-conditioned gates for sparse prompt activation and normalized on-the-fly fusion of current and historical task prompts. Therefore, GAP-Prompt differs from prior prompt integration methods not merely by replacing one weighting function with another, but by changing the role of weights from component composition or task-level insertion to instance-level routing and controlled historical knowledge fusion.

\begin{table}[h]
\centering
\caption{Comparison of GAP-Prompt with CODA-Prompt and RainbowPrompt}
\label{tab:formulation_comparison}
\resizebox{1.0\linewidth}{!}{
\begin{tabular}{l|ccc}
\toprule
\textbf{Aspect} & \textbf{CODA-Prompt~\citep{smith2023coda}} & \textbf{RainbowPrompt~\citep{hong2025rainbowprompt}} & \textbf{GAP-Prompt} \\
\midrule
Weight/gate type 
& Input-conditioned component weight 
& Task-conditioned insertion gate 
& Instance-conditioned activation gate \\
Weight/gate semantics 
& Component composition coefficient 
& prompt insertion decision
& prompt activation strength \\
Historical knowledge use 
& Frozen accumulated components 
& Evolved accumulated prompts 
& Selective current/history fusion \\
Inference behavior 
& Dense component assembly 
& Task-level evolved prompts 
& On-the-fly input-specific fusion \\
Sparsity 
& Dense weighting 
& Task-level selection 
& Thresholded sparse activation \\
\bottomrule
\end{tabular}
}
\end{table}

% \section{Additional Results}
% \subsection{Prompt Diversity}

% \subsection{Training Time Comparison}

% \subsection{Performance Across Varying Intra-task Diversity}

% \subsection{Additional Ablation Study}

% \subsection{Additional Results on Stanford Cars}

% \subsection{Additional Results in Domain Incremental Learning Setting}

% \section{Instance-Conditioned Gating (ICG) as Plugin}

% \section{Visualization}

% ssss

% \section{Efficiency Comparison with Task-base Adaptive Prompting} 

\section{Task querying accuracy and sensitivity analysis}
\label{app:task_matching_sensitivity}
At inference, GAP-Prompt first predicts a task index by matching the input query 
$\mathbf{q}$ with the learned task keys:
\begin{equation}
    \hat{t}(\mathbf{x})
    =
    \arg\max_{i\in\{1,\ldots,T\}}
    \gamma(\mathbf{q}(\mathbf{x}), \mathbf{k}^{i}),
\end{equation}
where $\gamma(\cdot,\cdot)$ denotes cosine similarity. The predicted task index 
$\hat{t}(\mathbf{x})$ determines the candidate prompt range used by Dynamic Knowledge Fusion (DKF). 
Since incorrect task querying may affect the final classification performance, we analyze both task querying accuracy and the sensitivity of GAP-Prompt to task routing errors.

\paragraph{Task querying accuracy.}
Let $\mathcal{D}_{\mathrm{test}}=\{(\mathbf{x}_n,y_n,t_n)\}_{n=1}^{N}$ denote the test set after learning all tasks, where $t_n$ is the ground-truth task index of sample $\mathbf{x}_n$. 
The task matching accuracy is defined as:
\begin{equation}
    \mathrm{Acc}_{\mathrm{query}}
    =
    \frac{1}{N}
    \sum_{n=1}^{N}
    \mathbb{I}\left[
        \hat{t}(\mathbf{x}_n)=t_n
    \right],
\end{equation}
where $\mathbb{I}[\cdot]$ is the indicator function, which equals $1$ if the condition is true and $0$ otherwise. This metric measures how often the task querying stage retrieves the correct task key before prompt fusion.

\begin{table}[h]
\centering
\caption{Task querying accuracy (\%) of the task-key retrieval module in GAP-Prompt after learning the final task under the 10-task class-IL setting. $\uparrow$ indicates that higher values are better}
\label{tab:task_querying_accuracy}

\resizebox{0.7\linewidth}{!}{
\begin{tabular}{lcc}
\toprule
\textbf{Dataset} & \textbf{Task querying accuracy} ($\uparrow$) & \textbf{Avg Acc} ($\uparrow$) \\
\midrule
CIFAR-100 & 55.75 & 89.24 \\
ImageNet-R & 44.69 & 78.72 \\
CUB-200 & 44.30 & 87.29 \\
\bottomrule
\end{tabular}
}
\end{table}

Tab.~\ref{tab:task_querying_accuracy} shows that the task querying accuracy is moderate across datasets, ranging from $44.30\%$ to $55.75\%$. 
Nevertheless, the final average accuracy remains high, especially on CIFAR-100 and CUB-200. 
This indicates that GAP-Prompt does not rely solely on perfectly correct task-key retrieval. 
Instead, because the gates are input-conditioned, GAP-Prompt can adapt prompt usage for each input even when the queried task index is imperfect. 
DKF can still reuse useful historical prompts within the candidate range, while ICG and thresholding help suppress weak or irrelevant prompt activations.

\paragraph{Classification accuracy under correct and incorrect task querying.}
To understand whether performance degradation mainly comes from incorrect task querying or from actual forgetting, we split the test samples into two groups according to the correctness of the queried task,
\begin{equation}
    \mathcal{S}_{\mathrm{correct}}
    =
    \left\{
    n \mid \hat{t}(\mathbf{x}_n)=t_n
    \right\},
    \quad
    \mathcal{S}_{\mathrm{wrong}}
    =
    \left\{
    n \mid \hat{t}(\mathbf{x}_n)\neq t_n
    \right\}.
\end{equation}
We then report classification accuracy separately for each group,
\begin{equation}
    \mathrm{Acc}_{\mathrm{cls}}^{\mathrm{correct}}
    =
    \frac{1}{|\mathcal{S}_{\mathrm{correct}}|}
    \sum_{n\in\mathcal{S}_{\mathrm{correct}}}
    \mathbb{I}
    \left[
        \hat{y}_n=y_n
    \right],
\end{equation}
\begin{equation}
    \mathrm{Acc}_{\mathrm{cls}}^{\mathrm{wrong}}
    =
    \frac{1}{|\mathcal{S}_{\mathrm{wrong}}|}
    \sum_{n\in\mathcal{S}_{\mathrm{wrong}}}
    \mathbb{I}
    \left[
        \hat{y}_n=y_n
    \right].
\end{equation}

This analysis helps distinguish the effect of incorrect task querying from class-level errors within the selected prompt space. 
Tab.~\ref{tab:task_querying_conditional_acc} further separates the effect of correct and incorrect task querying.
As expected, samples with correct task querying achieve higher classification accuracy across all datasets. 
However, the accuracy under wrong querying remains reasonably high, e.g., $81.91\%$ on CIFAR-100 and $82.57\%$ on CUB-200. 
This suggests that incorrect task querying does not necessarily lead to immediate failure, because the retrieved candidate range may still contain useful historical prompts and DKF can selectively fuse them according to instance-conditioned gates. 
The larger gap on ImageNet-R indicates that task querying errors are more harmful when the dataset contains stronger distribution shifts and more diverse visual styles.

\begin{table}[h]
\centering
\caption{Classification accuracy conditioned on task querying correctness. This analysis separates degradation due to incorrect task querying from degradation due to class-level forgetting or confusion}
\label{tab:task_querying_conditional_acc}
\begin{tabular}{lccc}
\toprule
\textbf{Dataset} & \textbf{Correct query} (\%, $\uparrow$) & \textbf{Wrong query} (\%, $\uparrow$) & \textbf{Avg Acc} (\%, $\uparrow$) \\
\midrule
CIFAR-100 & 94.35 & 81.91 & 89.24 \\
ImageNet-R & 90.65 & 68.99 & 78.72 \\
CUB-200 & 92.75 & 82.57 & 87.29 \\
\bottomrule
\end{tabular}
\end{table}

\paragraph{Over-estimation and under-estimation analysis.}
We further analyze the direction of task querying errors by defining:
\begin{equation}
    \Delta t_n
    =
    \hat{t}(\mathbf{x}_n)-t_n.
\end{equation}
When $\Delta t_n=0$, task querying is correct. When $\Delta t_n>0$, the queried task is later than the ground-truth task. In this case, the candidate prompt range still contains the correct historical prompts but also includes prompts from later tasks. ICG can mitigate this error by assigning low gates to irrelevant later-task prompts, which are further suppressed by thresholding. When $\Delta t_n<0$, the queried task is earlier than the ground-truth task. In this case, the correct task prompt may be excluded from the fusion range, making the error more harmful because DKF cannot directly access the prompt learned for the true task.

\begin{table}[h]
\centering
\caption{Sensitivity to the direction of task querying errors. $\Delta t=0$ indicates correct task querying, $\Delta t>0$ indicates over-estimation, and $\Delta t<0$ indicates under-estimation. All values are reported in percentage (\%).}
\label{tab:task_querying_error_direction}
\begin{tabular}{lcccc}
\toprule
\textbf{Dataset} & $\Delta t=0$ ($\uparrow$) & $\Delta t>0$ ($\uparrow$) & $\Delta t<0$ ($\uparrow$) & \textbf{Avg Acc} ($\uparrow$) \\
\midrule
CIFAR-100 & 94.35 & 82.90 & 83.50 & 89.24 \\
ImageNet-R & 90.65 & 71.41 & 66.38 & 78.72 \\
CUB-200 & 92.75 & 84.32 & 81.20 & 87.29 \\
\bottomrule
\end{tabular}
\end{table}

Tab.~\ref{tab:task_querying_error_direction} shows that both over-estimation and under-estimation reduce classification accuracy compared with correct task querying. 
However, their effects are not symmetric. 
On ImageNet-R and CUB-200, under-estimation is more harmful than over-estimation, with larger drops from $90.65\%$ to $66.38\%$ on ImageNet-R and from $92.75\%$ to $81.20\%$ on CUB-200. 
This is consistent with our inference design: when $\Delta t_n<0$, the correct task prompt may be outside the candidate range and cannot participate in DKF. 
By contrast, when $\Delta t_n>0$, the correct prompt remains within the range, while irrelevant later-task prompts can be suppressed by input-conditioned gates and thresholding. 
This analysis further supports the role of ICG and DKF as a second-stage correction mechanism after task querying.

% ========= ORACLE TASK QUERYING vs. All-task-fusion ===============
% ==============================
% \input{appendix/oracle_task_querying}

% If the gap between predicted querying and oracle querying is small, task-key retrieval is not the main bottleneck. If all-task fusion performs worse than predicted querying, this indicates that unrestricted global fusion introduces irrelevant prompt interference. If all-task fusion with threshold recovers performance, it suggests that ICG can suppress irrelevant prompts even without task-key range restriction, although at higher computational cost.

\section{Inference computational overhead and scalability}
\label{app:inference_efficiency}
A potential concern is that GAP-Prompt may introduce additional inference cost because DKF fuses current and historical prompts. 
We therefore analyze the additional operations beyond the frozen ViT forward pass and report latency across different task increments.

\paragraph{Additional inference operations.}
For an input image $\mathbf{x}$ after learning $t$ tasks, GAP-Prompt first extracts the query $\mathbf{q}(\mathbf{x})$ from the frozen ViT. 
The additional inference cost comes from three lightweight operations:
\begin{enumerate}
    \item \textbf{Task querying}: compute cosine similarities between $\mathbf{q}(\mathbf{x})$ and the learned task keys $\{\mathbf{k}^i\}_{i=1}^{t}$.
    \item \textbf{Instance-conditioned gating (ICG)}: compute gates $g_l^i(\mathbf{x})$ for candidate task prompts.
    \item \textbf{Dynamic knowledge fusion (DKF)}: aggregate only the candidate prompts whose gates remain active after thresholding.
\end{enumerate}
The ViT backbone is unchanged and remains frozen. Therefore, the overhead of GAP-Prompt is determined only by task-key querying, linear gating modules, and prompt fusion.

\paragraph{Complexity analysis.}
Let $d$ be the embedding dimension, $L_g$ the number of gated layers, $N_p$ the task-specific prompt length, and $\hat{t}(\mathbf{x})$ the queried task index. 
\textit{Task querying} requires:
\begin{equation}
    \mathrm{Cost}_{\mathrm{task-querying}} = \mathcal{O}(td),
\end{equation}
which is small compared with the ViT forward pass. 
Under the standard predicted-querying setting, GAP-Prompt does not evaluate all learned tasks for fusion. 
Instead, it uses the candidate range $\{1,\ldots,\hat{t}(\mathbf{x})\}$:
\begin{equation}
    \mathrm{Cost}_{\mathrm{ICG}}
    =
    \mathcal{O}\big(\hat{t}(\mathbf{x})L_gd\big).
\end{equation}

After applying the inference threshold $\zeta$, the number of active prompts can vary across layers and inputs. 
For an input $\mathbf{x}_n$ at the $l$-th MSA layer, let
\begin{equation}
    N_{\mathrm{active}}^{l}(\mathbf{x}_n)
    =
    \sum_{i=1}^{\hat{t}(\mathbf{x}_n)}
    \mathbb{I}\left[
        g_l^i(\mathbf{x}_n)>\zeta
    \right]
\end{equation}
denote the number of active task prompts at layer $l$. 
The fusion cost for this input is then:
\begin{equation}
    \mathrm{Cost}_{\mathrm{DKF}}(\mathbf{x}_n)
    =
    \mathcal{O}
    \left(
    N_p d
    \sum_{l=1}^{L_g}
    N_{\mathrm{active}}^{l}(\mathbf{x}_n)
    \right).
\end{equation}

\paragraph{Candidate gates before thresholding.}
Since task querying is input-dependent, each sample may have a different queried task index $\hat{t}(\mathbf{x}_n)$ and therefore a different candidate prompt range. 
For an input $\mathbf{x}_n$, the number of candidate task-layer gates before thresholding is:
\begin{equation}
    N_{\mathrm{cand}}(\mathbf{x}_n)
    =
    \hat{t}(\mathbf{x}_n)L_g,
\end{equation}
where $L_g$ is the number of gated MSA layers. 
We report the average number of candidate gates over the test set:
\begin{equation}
    \bar{N}_{\mathrm{cand}}
    =
    \frac{1}{N}
    \sum_{n=1}^{N}
    \hat{t}(\mathbf{x}_n)L_g.
\end{equation}

\paragraph{Effective active prompts after thresholding.}
After applying the inference threshold $\zeta$, the number of active gates can vary across layers and inputs. 
We therefore report the average number of active \emph{task-layer} prompts:

\begin{equation}
    \bar{N}_{\mathrm{active}}
    =
    \frac{1}{N}
    \sum_{n=1}^{N}
    \sum_{l=1}^{L_g}
    N_{\mathrm{active}}^{l}(\mathbf{x}_n)
    =
    \frac{1}{N}
    \sum_{n=1}^{N}
    \sum_{l=1}^{L_g}
    \sum_{i=1}^{\hat{t}(\mathbf{x}_n)}
    \mathbb{I}
    \left[
        g_l^i(\mathbf{x}_n)>\zeta
    \right].
\end{equation}
where $\mathbb{I}[\cdot]$ is the indicator function. 
This metric reflects the average number of task-specific prompts that actually participate in DKF across all gated layers for each input, instead of the worst-case cost of densely aggregating all stored prompts.

We also report the active ratio:
\begin{equation}
    r_{\mathrm{active}}
    =
    \frac{
    \bar{N}_{\mathrm{active}}
    }{
    \frac{1}{N}
    \sum_{n=1}^{N}
    \hat{t}(\mathbf{x}_n)L_g
    }.
\end{equation}

This metric quantifies how much thresholding reduces the effective fusion workload compared with using all candidate task-layer prompts. 

Tab.~\ref{tab:active_gates} shows that thresholding consistently reduces the effective number of prompts used by DKF. 
After learning 10 tasks, GAP-Prompt activates only $29.61$ out of $60.3$ candidate gates on CIFAR-100. 
For ImageNet-R and CUB-200, the active gates are $19.18/49.9$ and $7.06/52.7$, respectively. 
This confirms that DKF does not densely aggregate all candidate task-layer prompts; instead, it fuses only a thresholded subset of relevant prompts, reducing the effective inference workload as tasks accumulate.

\begin{table}[h]
\centering
\caption{Average number of active task-layer gates after inference thresholding. Candidate and active gates are averaged over all test samples. 
Active gates and active ratio indicate the effective sparsity of DKF after thresholding}
\label{tab:active_gates}
\setlength{\tabcolsep}{1.5mm}
\resizebox{0.9\linewidth}{!}{
\begin{tabular}{lcccc}
\toprule
\textbf{Dataset} & \textbf{Tasks learned} & \textbf{Avg candidate gates} & \textbf{Active gates} & \textbf{Active ratio} \\
\midrule
CIFAR-100 & 2  & 14.4 & 5.39 & 37.5\% \\
CIFAR-100 & 5  & 33.1 & 17.03 & 51.5\% \\
CIFAR-100 & 10 & 60.3 & 29.61 & 49.1\% \\
\midrule
ImageNet-R & 2  & 14.2 & 9.94 & 70.2\% \\
ImageNet-R & 5  & 29.7 & 17.08 & 57.4\% \\
ImageNet-R & 10 & 49.9 & 19.18 & 38.5\% \\
\midrule
CUB-200 & 2  & 15.3 & 3.08 & 20.1\% \\
CUB-200 & 5  & 34.0 & 5.50 & 16.2\% \\
CUB-200 & 10 & 52.7 & 7.06 & 13.4\% \\
\bottomrule
\end{tabular}
}
\end{table}

\paragraph{Latency across task increments.}
We measure inference latency after learning different numbers of tasks. 
All measurements are conducted under the same hardware, batch size, and input resolution. 
We report the latency of the frozen ViT backbone, the total latency of GAP-Prompt, and the additional overhead:
\begin{equation}
    \Delta_{\mathrm{latency}}
    =
    \mathrm{Latency}_{\mathrm{GAP}}
    -
    \mathrm{Latency}_{\mathrm{ViT}},
    \quad
    r_{\mathrm{overhead}}
    =
    \frac{
    \Delta_{\mathrm{latency}}
    }{
    \mathrm{Latency}_{\mathrm{ViT}}
    }.
\end{equation}

Tab.~\ref{tab:inference_latency} shows that the latency of GAP-Prompt increases gradually as more tasks are learned, which is expected because the candidate prompt range expands with the queried task index. 
However, the increase remains controlled across datasets. 
For example, on CIFAR-100, the latency increases from $19.87$ ms/image after 2 tasks to $30.77$ ms/image after 10 tasks. 
A similar trend is observed on ImageNet-R and CUB-200, where the latency after 10 tasks remains around $29$ to $30$ ms/image. 
This indicates that the additional cost introduced by task querying, ICG, and DKF grows moderately with the task sequence length, rather than causing an uncontrolled computational bottleneck.

\begin{table}[h]
\centering
\caption{Inference latency across task increments (ms/image). ``Overhead'' denotes the additional inference time over the frozen ViT backbone. $\downarrow$ indicates that lower values are better.}
\label{tab:inference_latency}
\setlength{\tabcolsep}{1.5mm}
\resizebox{\linewidth}{!}{
\begin{tabular}{lccccc}
\toprule
\textbf{Dataset} & \textbf{Tasks learned} & \textbf{ViT only} ($\downarrow$) & \textbf{GAP-Prompt} ($\downarrow$) & \textbf{Overhead} ($\downarrow$) & \textbf{Overhead ratio} ($\downarrow$) \\
\midrule
CIFAR-100 & 2  & 6.62ms & 19.87ms & 13.25ms & 200.0\% \\
CIFAR-100 & 5  & 6.62ms & 24.17ms & 17.55ms & 265.1\% \\
CIFAR-100 & 10 & 6.62ms & 30.77ms & 24.15ms & 364.9\% \\
\midrule
ImageNet-R & 2  & 6.61ms & 19.59ms & 12.98ms & 196.4\% \\
ImageNet-R & 5  & 6.62ms & 23.71ms & 17.09ms & 258.1\% \\
ImageNet-R & 10 & 6.64ms & 28.86ms & 22.22ms & 334.9\% \\
\midrule
CUB-200 & 2  & 6.60ms & 20.15ms & 13.55ms & 205.3\% \\
CUB-200 & 5  & 6.61ms & 25.37ms & 18.76ms & 283.9\% \\
CUB-200 & 10 & 6.63ms & 30.38ms & 23.76ms & 358.4\% \\
\bottomrule
\end{tabular}
}
\end{table}

\paragraph{Comparison with prompt-based CL methods.}
We further compare inference behavior with representative prompt-based CL methods. 
DualPrompt~\citep{wang2022dualprompt} uses a fixed set of task prompts once the task is retrieved. CODA-Prompt~\citep{smith2023coda} performs dense component assembly over its accumulated prompt-component dictionary. 
RainbowPrompt uses evolved prompts and task-level gates. 
In contrast, GAP-Prompt uses task querying to restrict the candidate prompt range to $1:\hat{t}(\mathbf{x})$, and further applies thresholded instance-conditioned gates to remove weak prompt activations before DKF. 
As a result, GAP-Prompt avoids dense aggregation over all stored prompt components and uses fewer effective prompts during inference.

Since GAP-Prompt performs input-dependent routing, its GFLOPs can vary across samples. 
We therefore report the average measured GFLOPs over the test set. 
As shown in Tab.~\ref{tab:inference_efficiency_comparison}, GAP-Prompt achieves comparable GFLOPs to DualPrompt, CODA-Prompt, and RainbowPrompt, while using substantially fewer extra stored parameters than CODA-Prompt and RainbowPrompt on CIFAR-100. Regarding extra inference parameters, RainbowPrompt requires 1.85M parameters after inference-time discard, while GAP-Prompt consistently uses only 1.36M.
The similar GFLOPs are expected because the frozen ViT backbone dominates the overall computation. 
However, GAP-Prompt is more efficient in terms of prompt parameterization and effective prompt usage: it relies on compact task-specific prompts, lightweight linear gates, and thresholded DKF, rather than dense component aggregation or additional prompt-evolving modules. 
This demonstrates that GAP-Prompt achieves adaptive prompt reuse with comparable inference GFLOPs, fewer extra stored/inference parameters, and a more selective prompt-fusion process.

\begin{table}[h]
\centering
\caption{Inference efficiency comparison with prompt-based CL methods on CIFAR-100 under the 10-task class-IL setting. GFLOPs are measured under the same hardware and batch size, and averaged over the test set for GAP-Prompt due to input-dependent routing and thresholded DKF.}
\label{tab:inference_efficiency_comparison}
\setlength{\tabcolsep}{1.2mm}
\resizebox{1.0\linewidth}{!}{
\begin{tabular}{lccccc}
\toprule
\textbf{Method} & \textbf{Extra stored params} & \textbf{Extra inference params} & \textbf{Candidate set at inference} & \textbf{Active prompts/layers} & \textbf{GFLOPs} ($\downarrow$) \\
\midrule
ViT only & 0M & 0M & -- & -- & 33.70 \\
\midrule
DualPrompt & 1.10M & 1.10M & Retrieved task prompt & Fixed & 67.44  \\
CODA-Prompt & 3.99M & 3.99M & Accumulated components & Dense component assembly & 67.44 \\
RainbowPrompt & 6.44M & 1.85M & Evolved task prompts & Task-level gates & 67.46  \\
GAP-Prompt & 1.36M & 1.36M & $1:\hat{t}(\mathbf{x})$ & Thresholded instance gates & 67.39 \\
\bottomrule
\end{tabular}
}
\end{table}

\paragraph{Discussion.}
Overall, GAP-Prompt remains inference-efficient by restricting prompt computation rather than densely using all stored prompts. Task querying limits the candidate range to $1:\hat{t}(\mathbf{x})$, while thresholding removes weak gates before DKF. As a result, the practical fusion cost depends on the active task-layer prompts after thresholding, instead of the total number of stored historical prompts. 
Together with the lightweight linear ICG module, this design keeps the inference overhead controlled as the number of learned tasks increases.

\input{appendix/visualization}

% As shown in Fig.~\ref{fig:gate_visualization}, different inputs produce different activation patterns across tasks and layers, even when they belong to the same task. 
% This supports our claim that ICG does not learn a task-shared prompting topology, but instead generates instance-conditioned prompt activation strengths.

% We also visualize the thresholded gates used during inference:
% \begin{equation}
%     \tilde{g}_l^i(\mathbf{x})
%     =
%     g_l^i(\mathbf{x})
%     \cdot
%     \mathbb{I}
%     \left[
%         g_l^i(\mathbf{x})>\zeta
%     \right].
% \end{equation}
% The thresholded maps show that weak gates are suppressed, resulting in sparse task-layer prompt routing. 
% This provides qualitative evidence that DKF does not densely aggregate all stored prompts, but selectively fuses only prompts that are relevant to each input.

%% file: appendix/visualization.tex
% \section{Visualization of instance-conditioned gating}
% \label{app:visualization}
% To qualitatively verify whether instance-conditioned gating (ICG) learns input-specific prompt routing, we visualize the gate values produced for different test images. 
% For each input $\mathbf{x}$, we construct a gate matrix:
% \begin{equation}
%     \mathbf{G}(\mathbf{x})
%     =
%     \left[
%     g_l^i(\mathbf{x})
%     \right]_{i=1,\ldots,\hat{t}(\mathbf{x}),\; l=1,\ldots,L_g},
% \end{equation}
% where rows correspond to task-specific prompts and columns correspond to gated MSA layers. 

% ...